%% file: wrapper.tex
\documentclass[submit]{bioRxiv}
\usepackage[numbers]{natbib}

\usepackage{graphicx}  
\usepackage{textcomp} 
\usepackage{gensymb}  
\usepackage{siunitx}  
\usepackage{amsmath} 
\usepackage{tabularx}  
\usepackage{longtable}  
\usepackage{hyperref}  
\usepackage{lipsum} 
\usepackage{parskip} 
\usepackage{mathtools}
\usepackage{algorithm}
\usepackage{float}

\newcommand{\x}{\mathbf{x}}
\renewcommand{\i}{\mathbf{i}}
\renewcommand{\j}{\mathbf{j}}

\renewcommand{\P}{\mathcal{P}}
\newcommand{\J}{\mathcal{J}}
\newcommand{\X}{\mathcal{X}}
\renewcommand{\L}{\mathcal{L}}

\newcommand{\D}{\Delta}

\newcommand{\TMP}{\textbf{TMP}}
\newcommand{\TMPH}{\textbf{TMP}^{\textbf{H}}}

\newcommand{\FB}{\textbf{FB}}
\newcommand{\Th}{\Theta}
\renewcommand{\th}{\theta}
\newcommand{\I}{\mathcal{I}}
\newcommand{\E}{\mathbb{E}}

\DeclareMathOperator*{\argmin}{arg\,min}

\begin{document}

\include{main}

\include{supplement}

\end{document}

%% file: main.tex
\leadauthor{Bear}
\title{Unifying (Machine) Vision via Counterfactual World Modeling}
\shorttitle{Counterfactual World Modeling}

\author[1]{Daniel M. Bear}
\author[2]{Kevin Feigelis}
\author[3]{Honglin Chen}
\author[4]{Wanhee Lee}
\author[3]{Rahul Venkatesh}
\author[3]{Klemen Kotar}
\author[5]{Alex Durango}
\author[1,3,6]{Daniel L. K. Yamins}

\affil[1]{Department of Psychology, Stanford University, Stanford, CA 94305}
\affil[2]{Department of Physics, Stanford University, Stanford, CA 94305}
\affil[3]{Department of Computer Science, Stanford University, Stanford, CA 94305}
\affil[4]{Department of Applied Physics, Stanford University, Stanford, CA 94305}
\affil[5]{Neurosciences Graduate Program, Stanford University, Stanford, CA 94305}
\affil[6]{Wu Tsai Neurosciences Institute, Stanford University, Stanford, CA 94305}

\date{May 2023}

\maketitle

\begin{abstract}
\input{abstract}
\end{abstract}


\begin{corrauthor}
dyamins\at gmail.com
\end{corrauthor}

\section{Introduction}\label{sec:introduction}
\input{introduction}

\section{Masked Predictors} \label{sec:masked_predictors}
\input{masked_predictors}

\section{Tasks as Counterfactual Derivatives of $\Psi$} 
\label{sec:tasks}
The observations of the previous section show how the CWM, $\Psi$, has implicitly learned powerful information about the physical structure of scenes and factors the basic causes of change across frames.
The information about the second frame that comes in the form of a few second-frame patches is similar to ``action'' data, describing in an action-space-agnostic way a minimal amount of visual information necessary for $\Psi$ to identify state changes. 
Steerability with a small number of tokens makes the CWM a possible interface for many visual tasks: different computations can be specified by feeding $\Psi$ different counterfactual ``prompts''. 
In this section we show how several complementary visual properties can be conceptually unified: optical flow, occlusion, movable-object segmentation, and relative depth can all be extracted from the CWM through variants of the same prompting procedure.

\subsection{Optical Flow}\label{sec:flow}
\input{flow}

\subsection{Movable-Object Segmentation}\label{sec:segmentation}
\input{segmentation}

\subsection{Relative Depth}\label{sec:depth}
\input{depth}

\section{Discussion}\label{sec:discussion}
\input{discussion}

\bibliographystyle{unsrtnat}
\bibliography{wrapper.bbl}

%% file: abstract.tex
Leading approaches in machine vision employ different architectures for different tasks, trained on costly task-specific labeled datasets.
This complexity has held back progress in areas, such as robotics, where robust task-general perception remains a bottleneck.
In contrast, "foundation models" of natural language have shown how large pre-trained neural networks can provide zero-shot solutions to a broad spectrum of apparently distinct tasks.  
Here we introduce Counterfactual World Modeling (CWM), a framework for constructing a visual foundation model: a unified, unsupervised network that can be prompted to perform a wide variety of visual computations.
CWM has two key components, which resolve the core issues that have hindered application of the foundation model concept to vision.
The first is structured masking, a generalization of masked prediction methods that encourages a prediction model to capture the low-dimensional structure in visual data.
The model thereby factors the key physical components of a scene and exposes an interface to them via small sets of visual tokens.
This in turn enables CWM's second main idea -- counterfactual prompting -- the observation that many apparently distinct visual representations can be computed, in a zero-shot manner, by comparing the prediction model's output on real inputs versus slightly modified ("counterfactual") inputs.
We show that CWM generates high-quality readouts on real-world images and videos for a diversity of tasks, including estimation of keypoints, optical flow, occlusions, object segments, and relative depth. 
Taken together, our results show that CWM is a promising path to unifying the manifold strands of machine vision in a conceptually simple foundation.

%% file: introduction.tex
Within their first years of life, humans acquire an unparalleled ability to understand visual scenes and make predictions about them: what objects are present, how they are arranged in space, where they will (or could) move, and how they might interact.
These abilities in turn support a panoply of complex behaviors, such as object manipulation and tool use, long-range physical prediction and planning, spatial navigation, language grounding, and social interaction. 

Despite great recent progress in modeling high-level cognitive functions, such as language understanding and generation, even today's large-scale AI algorithms fall far short of human-like visual scene-understanding.
This has held back artificial systems, such as robots, that would profitably use such abilities for model-predictive control and planning.
Importantly, because they arise so early in human development -- and are shared in part by other animals, like non-human primates -- visual scene understanding and world-modeling abilities almost certainly do not require language learning or supervision.
This suggests that bridging the machine-human gap in vision (or pre-linguistic sensorimotor abilities more broadly) may require more powerful and efficient new forms of unsupervised learning.

One appealing approach, inspired by vision science and visual systems neuroscience, is to break the problem of scene understanding down into sub-tasks that can be solved with special-purpose modules.
The human visual system extracts a wide range of low- and mid-level representations of scenes, including (among many others) contours and border ownership~\cite{zhou2000coding}, optical flow and self-induced motion~\cite{beauchemin1995computation}, perceptual groups and figure-ground segmentations~\cite{lamme1995neurophysiology, treisman1982perceptual}, a ``2.5D sketch'' of depth and surface normal vectors~\cite{marr2010vision}, and estimates of 3D object shape~\cite{todd2004visual}.
Individual computer vision algorithms successfully model some of these ``intermediate'' computations, but most appear to be well below human level and fail to capture more complex visual behaviors~\cite{bear2021physion}, raising deep questions about this agenda: 
which intermediates are necessary and sufficient for holistic scene understanding?
How can better algorithms be found without hard-to-obtain labeled data?\footnote{...which humans and animals do without. While supervised algorithms for (e.g.) object categorization or segmentation can succeed in producing partially accurate proxies for adult visual behaviors~\cite{rajalingham2018large} and their neural correlates~\cite{yamins2014performance}, these approaches are implausible as models of real (i.e. unsupervised) visual learning~\cite{zhuang2021unsupervised}.}
Are improvements in one module useful for improving others?
Do the many intermediates need to be assembled carefully into a unified architecture, or can a generic procedure discover automatically how to make use of them?

A promising alternative may dispense with these challenges.
Recently, the ``foundation model'' approach~\cite{bommasani2021opportunities} has driven rapid progress in natural language generation and understanding tasks, without breaking them down into pieces that require special methods.
At a high level, this approach has two key components:
\begin{itemize}
  \item A \emph{pretrained model}, typically a large, relatively unspecialized neural network, trained to solve a ``masked token prediction'' task on a large real-world dataset; and, 
  \item A \emph{generic task interface}, a simple process for translating any task within a wide domain into a zero- (or few-)shot input to the pretrained model.  
\end{itemize}

\begin{figure}[t]
\centering
\includegraphics[width=\linewidth]{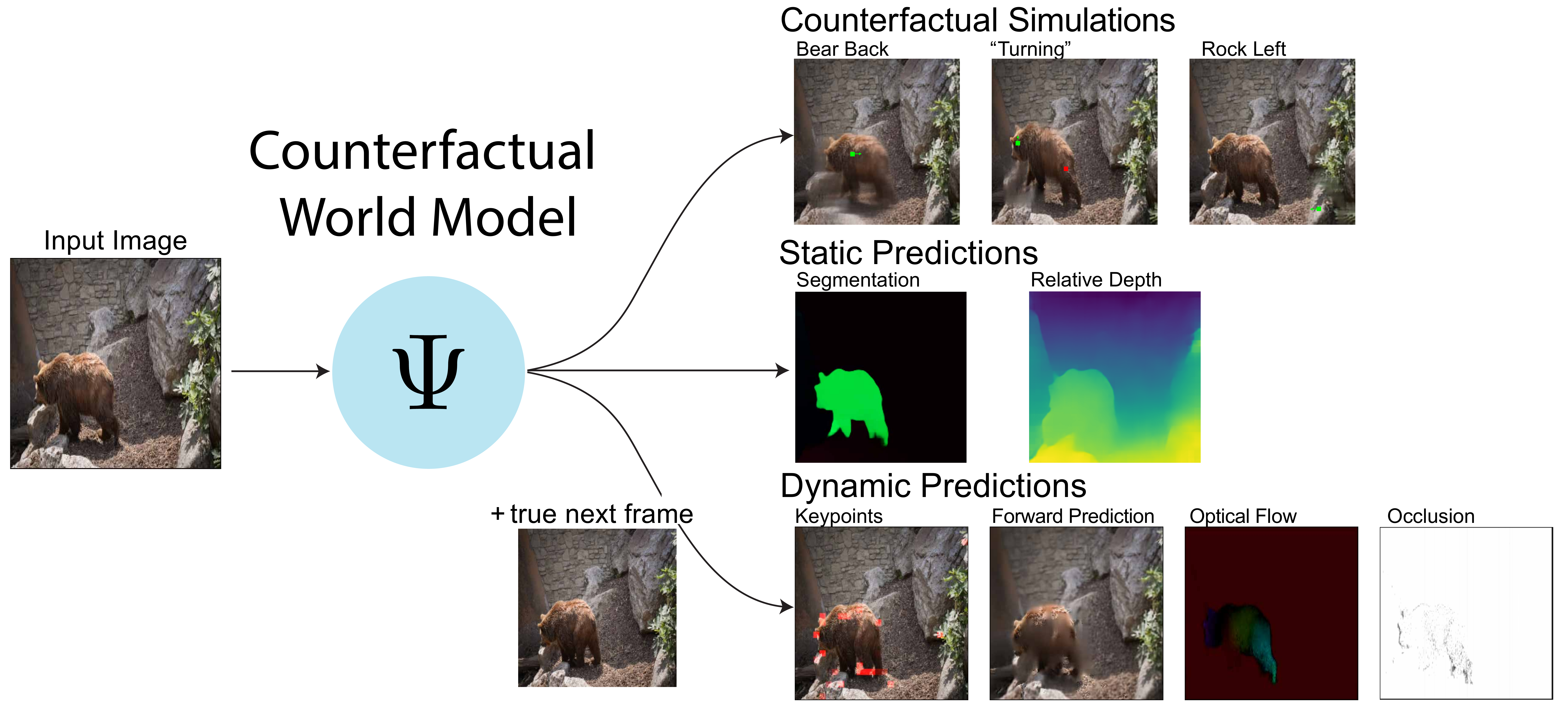}
\caption{
    \textbf{Counterfactual World Modeling unifies visual computations.}
    A Counterfactual World Model, $\Psi$, takes an input image (or video) and constructs an internal representation of the scene. This representation can be used to generate \textit{counterfactual simulations} of what would happen if different events occurred (\textbf{top right}) and to estimate properties of the real scene (\textit{static and dynamic predictions}, \textbf{middle and bottom right}). Running counterfactual simulations requires input to $\Psi$ in the form of ``counterfactual prompts,'' which indicate which image patches are moving (green squares with arrows) and which are static (red squares.) Predictions about a scene's dynamical properties (keypoints, forward predictions, optical flow, and occlusions) require a second frame of the true input video.
}
\label{fig:unified}
\end{figure}

In the case of language, simple and powerful versions of both of these ideas have been discovered. 
Effective large language models, like GPT-n~\cite{radford2019language}, are trained on large corpuses of web text to take in a context string and predict the likely next word. 
The next-word prediction task is straightforward to formulate (if not to solve) and remains unsurpassed, despite attempts to identify more powerful training methods.
Moreover, the form of this training immediately yields a universal interface: 
a well-trained model prompted with a natural language description of a task -- that is, provided with this description as its context string -- \textit{must} provide an attempted solution, its prediction of what text would naturally follow the prompt.
Thus language foundation models have gained widespread success from their remarkable simplicity, compared to the wiring together of specialized subsystems that predominates in computer vision.

But applying the foundation model framework to vision has not been as straightforward.
Both key ingredients are harder to formulate and successfully implement: 
masked prediction in vision, in the guise of future prediction of video frames~\cite{voleti2022masked, yan2022temporally} (or meaningful image-derived latents~\cite{nair2022r3m}) from previous context frames (or latents), has remained a largely unsolved problem~\cite{bear2021physion};
and it is not at all clear how to elicit most of the ``intermediate'' computations above through an interface to a single vision model.
For instance, what sort of input -- which must be visual in nature, if training were purely on unlabeled visual data -- would cause a vision model to output object segments?
Recent unified vision models emphasize a shared architectural stem from which many outputs are produced, but they all require some combination of (i) supervised training on labeled datasets for each supported task, (ii) task-specific decoder architectures, and (iii) a language prompting interface (and therefore training on joint visual-textual data at some stage)~\cite{jaegle2021perceiver,lu2022unified,chen2022unified,wu2023visual,li2023uni,wang2023images}.
They do not explain, therefore, how to derive a task-general interface from a single unsupervised training process, which we take to be the kernel of the foundation model concept.

Here, we propose to resolve these issues with an approach we call \emph{Counterfactual World Modeling (CWM)}.  
The CWM framework has two basic concepts, which build on the two fundamental components of foundation models:
\begin{enumerate}
    \item \textbf{[Pretraining via structured masked prediction.]} The recent Masked Autoencoder models~\cite{he2022masked} directly translate the masked prediction task from text to visual data (images and video); however, these models do not appear to learn representations as general as language models.
    In stark contrast to the language domain, we find that simple but nontrivial variations of the masks used to train vision models produce highly distinct learned representations. 
    In particular, we show that a \emph{temporally-factored} masking structure teaches models to separate \emph{appearance} information from \emph{dynamics} information. 
    The resulting prediction model encodes powerful implicit information about the physical structure of scenes, objects, and their dynamics.
    \item \textbf{[Generic task interface via counterfactual prompting.]} Different visual representations do not, at first blush, appear to have much in common: state-of-the-art computer vision uses quite distinct methods to extract, e.g., keypoints, optical flow, and object segments, with few obvious relationships between them. Yet we show that many such concepts arise from the above prediction model via \textit{counterfactual prompting}: by systematically modifying visual inputs in simple ways and inspecting the perturbed model responses, core computer vision concepts can be extracted in a zero-shot manner. For example, extracting optical flow is equivalent to asking for a prediction of where a hypothetical visual marker would move from one frame to the next. Moreover, we show that these counterfactuals can be mathematically formalized as \textit{derivatives} of the underlying prediction model itself, which provides a way both to identify ``natural'' counterfactual prompts and to implement them efficiently with autograd methods. Figure \ref{fig:unified} illustrates the use of a Counterfactual World Model, $\Psi$, in both generating counterfactual simulations and predicting properties of the true visual input.
\end{enumerate}
In this work we show how basic application of Counterfactual World Modeling can construct a variety of visual concepts, including keypoints, optical flow and occlusion, object segmentation, and monocular depth. 
Several of these constructions involve ``bootstrapping,'' using one extracted representation to help extract another.
Bootstrapping is simple in the CWM framework because it is effectively an application of the chain rule; for example, we illustrate below how object segmentation can be considered the derivative of a model that estimates optical flow, which in turn is the derivative of a temporally-factored masked prediction model.

Our results illustrate how a high-performing, ``true'' foundation model can be formalized and trained on unlabeled visual data.
The CWM approach thus synthesizes a plethora of classical computer vision concepts within the modern foundation model framework.
Because CWM does not require supervision or specialized architectures for different sub-tasks, it offers a practical solution for many visual problems and -- for the first time -- a \textit{unified computational theory} of how these abilities could arise in a biological system.

%% file: masked_predictors.tex
Many classical forms of unsupervised learning can be generalized as a simple training task on unlabeled data, \textit{masked token prediction}.
Each element of a dataset (e.g. a piece of text, an image, a video, or an audio stream), is broken into non-overlapping chunks (``tokens''); during training, a portion of the tokens is randomly masked out and the parameters of a neural network are optimized to predict this missing information from the remaining ``visible'' (unmasked) tokens.
Intuitively, because the neural network model must infer things about the structure of the training data to do this task well, its learned representation may generalize to other downstream tasks that depend on that structure.

This intuition has been borne out dramatically in the case of Large Language Models (LLMs) and text data.
Highly capable and general language models can be trained on an especially simple version of masked token prediction in which only the final token (roughly, the last word) of a text stream must be predicted from prior context.
This model and training structure creates a natural, universal task interface: different tasks are turned into queries simply by feeding the task description, which is a text string, as input to the prediction model.

The Masked Autoencoder (MAE) framework\footnote{The MAE is unfortunately named, as nothing in an input image is autoencoded: the patches to be filled in are specifically \textit{not} part of the input to the model. For this reason, we adopt the term \textit{masked prediction} throughout.} is a direct analogue of LLMs' model architecture and masked token prediction training to visual data~\cite{he2022masked}. 
MAEs have gained widespread use because they are easy to scale and transfer to downstream tasks (via supervision) about as well as more complex unsupervised methods. 
The largest MAEs (and their natural extension to multi-frame videos, VMAEs) learn to capture some of the higher-level structure in visual data, such as the shapes, textures, and subpart arrangements of common objects.

However, (V)MAEs appear to lack LLMs' generality and power as both an unsupervised representation learning method and a universal task interface.
While (V)MAEs perform only slightly worse than alternative unsupervised learning frameworks, such as contrastive learning~\cite{chen2020simple,grill2020bootstrap}, the state-of-the-art methods for most vision tasks are almost exclusively supervised models with specialized architectures~\cite{kirillov2023segment}; this is emphatically different from language modeling, where foundation models are to a first approximation the best at \textit{all} language tasks.
Furthermore, as with all other unsupervised vision approaches, (V)MAEs are applied to various tasks through transfer learning.
This requires further training on separate labeled datasets, in contrast to LLMs' direct applicability to any language task via zero-shot prompting.

Why is it hard to fully translate LLMs' success into a vision foundation model?
We hypothesize that it is fundamentally because single text tokens -- roughly, words -- more strongly constrain the statistics of surrounding text than do single visual patch tokens.
Reading the word ``rhino'' dramatically narrows the sequence of words that can be expected to follow, whereas seeing a small patch of an image only very weakly constrains what objects are present in the image, where they are in 3D space, how they are moving, and so on.
This is why it is possible to use LLMs as a universal task interface: a small number of text tokens is often enough context to serve as a meaningful task prompt.
(V)MAEs, on the other hand, require many tokens (e.g. 10\% of an image) before they begin to generate plausible completions of visual data, making it very challenging to steer such models via prompting.

Yet there \textit{is} substantial, low-dimensional structure in visual data. 
Our key insight is that this structure can be captured and concentrated into a small number of visual tokens. 
This works by training masked prediction models with more structured forms of token masking (i.e., not uniformly sampling visual patches to be masked out).
In particular, we note that visual scene \textit{dynamics} are highly constrained on very short timescales, because they stem from a few physical objects moving in a relatively small number of ways.
A form of training we call \textit{temporally-factored masked prediction} causes a standard vision transformer network to learn this structure implicitly, which creates a natural set of ``handles'' with which visual data can be manipulated explicitly -- the basis for steering the model to do different visual tasks (detailed in \S\ref{sec:tasks}).
Together the procedures of (i) training a model with structured masking and (ii) prompting it with simple counterfactual inputs constitute Counterfactual World Modeling.

\begin{figure}[t]
\centering
\includegraphics[width=\linewidth]{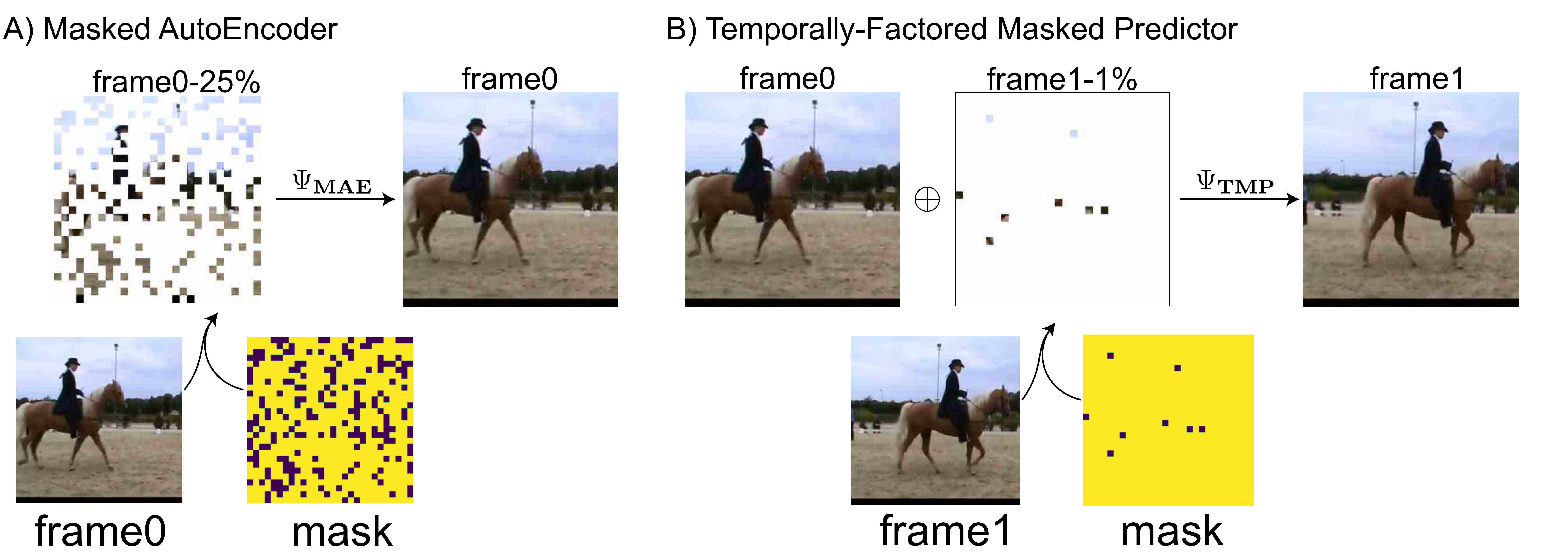}
\caption{
    \textbf{Masked Predictors.}
    \textbf{A)} The Masked Autoencoder (MAE) masking policy reveals an intermediate fraction ($\sim$ 25 \%) of patches in a frame. The standard Video MAE policy extends the MAE policy uniformly to video frames.
    \textbf{B)} The temporally-factored masked predictor policy asymmetrically reveals all the patches in \textbf{frame0} and a small fraction ($\sim$ 1\%) of patches in \textbf{frame1}.  This masking policy forces the resulting predictor $\Psi_{\textbf{TMP}}$ to factor appearance and dynamics. 
}
\label{fig:tmp}
\end{figure}

\begin{figure}[t]
\centering
\includegraphics[width=\linewidth]{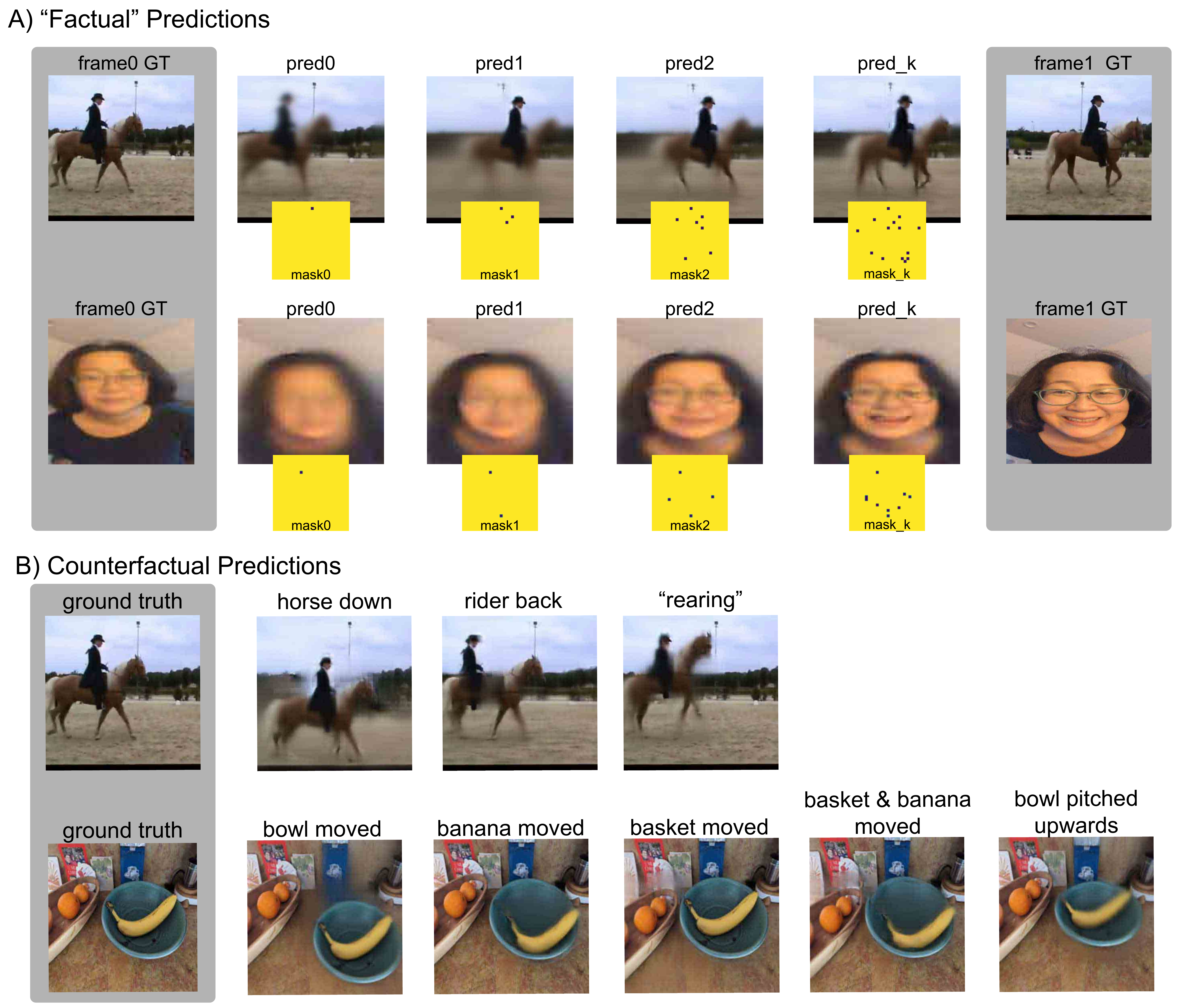}
\caption{
    \textbf{Factual and Counterfactual Predictions with the temporally-factored masked predictor $\Psi_{\textbf{TMP}}$.}  
    \textbf{A) Factual Predictions.} Sequence of predictions as more patches from frame 1 are revealed.  Top row: This case consists of two frames drawn 2.4s apart from a Kinetics 400 movie. Here, the first patch fixes the camera panning by revealing a little bit of the background. The second patch essentially sets the location of the foreground object (the horse and rider).  Subsequent patches fill in details of the object motion (e.g. horse leg positioning). Bottom row: This case consists of two frames drawn 2.2s apart from a home movie captured on a smartphone. Here, the first patch reveals that the person has moved forward and to the right in the frame. The subsequent several patches then fix the angle of the face, keeping the expression constant.  The last several patches reveal that the expression has changed to a smile.
    \textbf{B) Counterfactual Predictions.} $\Psi$ can simulate counterfactual motions from single frames. The input to the predictor is {\bf frame0} together with a simulated second frame in which one or several patches from {\bf frame0} have been moved to desired locations to seed motion (see Fig \ref{fig:seg1}a for illustration.) Choosing different patches to move and directions in which to move them generates different counterfactual next-frames.
}
\label{fig:tmp_predictions}
\end{figure}

\textbf{Factoring appearance from dynamics via structured masking.}
Masked prediction models may learn different things depending on how tokens of data are masked during training.
A \emph{masking policy} is a (random-valued) function $g$ that assigns to each potential input $\x$ a subset of token indices in $\x$ that will be \emph{visible}. The \emph{masked prediction model associated with $g$} is the function $\Psi_g$ that takes as input the visible tokens (those at indices $i$ for which $i \in g(\x)$) and estimates the values of invisible tokens (those at indices $i$ for which $\i \notin g(\x)$).\footnote{See supplementary section \ref{s_sec:formaldef} for a more precise mathematical definition.}

The standard masked autoencoder $\Psi_{\textbf{MAE}}$ is defined by the \textit{uniform} masking policy $g_p$ that, for some fixed fraction $p \in [0, 1]$, masks out uniformly sampled\footnote{Typically implemented by randomly permuting the patch indices and masking out the last $p$.} proportion $p$ of patches of the input (Fig. \ref{fig:tmp}a)~\cite{he2022masked}. The VMAE is given by extending this policy to all frames in a video or by uniformly sampling ``tubes'' of patches at the same location across multiple frames~\cite{tong2022videomae,feichtenhofer2022masked}. For the MAE, optimal values of $p$ for representation learning tend to occur at $p \sim 0.75$~\cite{he2022masked}. This is because setting $p$ too high provides too little information to train an effective prediction model, while setting $p$ too low provides \emph{too much} information and thus not enough incentive to learn general visual representations.

Here, we introduce an alternative masking scheme, based on the observation that on very short timescales ($\sim$250 ms) the next frame of a video can be predicted well simply by \textit{copying} large chunks of the previous frame to nearby locations;
the chunks correspond to independently moving objects, and object deformation and out-of-plane motion play a far smaller role than at longer timescales.
In other words, information about scene \textit{dynamics} has low-dimensional structure in this regime even if per-frame appearance does not.
This asymmetry between dynamics and appearance can be harnessed with an asymmetric masking policy.

Specifically, given an RGB frame pair $\x = (x_t, x_{t+\D})$, consider the policy $g_{q, p}$ in which the fraction $q$ of patches in frame $x_t$ are masked, while fraction $p$ of patches in $x_{t+\D}$ are masked. 
When $q \sim 0$ and $p \sim 1$ (we set $q = 0$ and $p = 0.99$), the model $\Psi_g$ must complete the second frame given only a few patches of it, plus the entire first frame (Fig \ref{fig:tmp}b);
the only way it can do this is by detecting how objects are moving from the small set of second-frame patches, then applying these transformations to the full objects in the first frame.
Thus a prediction model trained with this \textit{temporally-factored} masking policy, $\Psi_{\textbf{TMP}}$, will learn implicitly to factor information about scene state (e.g. object position, pose, size, and camera motion) from information about scene dynamics -- if it learns a general solution at all.
Moreover, the higher the second-frame masking fraction $p$, the more information about scene dynamics will have to be concentrated into a small number of visual tokens.

\textbf{Temporal factoring in ``factual'' predictions.}
We trained $\Psi_{\textbf{TMP}}$ on a real-world video dataset (Kinetics400~\cite{kay2017kinetics}) and evaluated its predictions when prompted with small sets of second-frame patches (see Supplementary section \ref{s_sec:tmp} for training details).
This revealed that $\Psi_{\textbf{TMP}}$ does in fact learn to factor appearance from dynamics. 
Going forward, we will drop the subscript and refer to this prediction model simply as $\Psi$.

The factoring properties of $\Psi$ can be seen when performing a series of increasingly accurate ``factual predictions'' of actual videos (Fig \ref{fig:tmp_predictions}a).
Given a frame pair $(x_t, x_{t + \D})$ taken from a real movie, we start with one patch in $x_{t+\D}$ given and observe how the prediction improves and changes as additional patches are revealed. As the figure illustrates, a single revealed patch allows $\Psi$ to capture large-scale changes in the scene, e.g. camera panning causing background motion, or gross object motion. 
As one or two more patches are added, additional, independent features of the true scene dynamics are well-predicted, including foreground object position and pose changes. As further patches are added, small details come into focus.
Throughout this process, position- and pose-invariant object appearances change very little (other than becoming less blurry), implying that ``prompting'' $\Psi$ with individual patches primarily conveys information about scene dynamics.

\textbf{Generating counterfactual predictions.} 
Because its representation of scene dynamics has been concentrated in a small number of tokens, $\Psi$ supports the easy generation of \emph{counterfactual predictions} (Fig. \ref{fig:tmp_predictions}b).

By definition, a \emph{counterfactual} is an input to the predictor that has been modified from the initial ground-truth input in some systematic way.
We use two basic flavors of counterfactual: ``appearance counterfactuals,'' which involve modifications to the first frame ($x_t$), and ``motion counterfactuals,'' which involve modifications to the second frame ($x_{t+1}$).
Given a counterfactual input, the associated \emph{counterfactual prediction} is the output of $\Psi$ on this made-up datum.

Figure \ref{fig:tmp_predictions}b shows counterfactual predictions for a series of motion counterfactuals. 
These are generated by starting with a single image as the first-frame input to $\Psi$ and producing counterfactual second-frame inputs by translating one or two patches in the original image by small amounts. 
The results show how it is possible to use a small number of simple counterfactual modifications to move semantically meaningful sub-components of the image (e.g. objects) in an independent fashion. For this reason, we will sometimes refer to $\Psi$ as a \emph{counterfactual world model} (CWM).

Section \S\ref{sec:tasks} shows how many seemingly distinct visual computations can be formulated as counterfactual comparisons: contrasts between $\Psi$'s predictions on real inputs versus counterfactual inputs.  
The CWM's ability to be flexibly prompted yields zero-shot solutions to a wide variety of tasks; this bears much closer resemblance to prompting LLMs with simple context strings, compared with standard MAEs or VMAEs needing many visible patches to produce realistic outputs.

\textbf{Extensions of the prediction model and masking policy.}
A natural extension of $\Psi$ offers a way to prompt the model with information about viewer motion, which can confound estimates of distal scene dynamics.
During training the prediction model can be conditioned both on visible patches of video and on data correlated with head motion, as several recent large-scale video datasets come equipped with camera accelerometry (IMU) data~\cite{grauman2022ego4d}.\footnote{These data are readily available to biological organisms from in-head ear canal gyroscopes~\cite{day2005vestibular}, proprioceptive feedback about relative head-body-angles, and motor efference copies~\cite{crapse2008corollary}}
The joint video-and-head-motion-conditioned masked predictor is denoted $\Psi^{\textbf{H}}$ (See supplementary section \ref{s_sec:head_motion} for details).
We show below that prompting this model with different \textit{counterfactual} head motions offers a way to improve object segmentation and estimate depth from a single image.
More generally, a data stream from any modality (including true actions, like head motions) can be combined with a visual masked prediction model to discover structure in their joint distribution.

Furthermore, training with different masking policies can cause masked predictors to learn about different structure in visual data.
These include policies that (i) blend TMP and MAE properties by partly masking the first frame, (ii) achieve better backward predictions by combining forward and reverse asymmetric masking, and (iii) perform extrapolation and interpolation of masked video frames. See Supplementary section \ref{s_sec:mask_variants} for further detail.

\textbf{Keypoints.} 
Keypoints are local features of a scene that capture \textit{more information} than other features~\cite{minderer2019unsupervised,mian2008keypoint}.
In our approach, keypoints are naturally defined as those tokens that, when made visible to the prediction model $\Psi$, yield lowest reconstruction error on the rest of the scene.
Keypoints turn out to be good loci for exerting counterfactual control.
Intuitively, this is because they are the few visual tokens most able to reduce uncertainty and yield a concrete prediction about many other tokens.
Note that, whereas the counterfactual prompting constructions below (see \ref{sec:tasks}) involve perturbing the visual input to $\Psi$, keypoints arise by varying the prediction model's \emph{input mask};
we therefore see them as part of the CWM approach.

\begin{figure}[t]
\centering
\includegraphics[width=\linewidth]{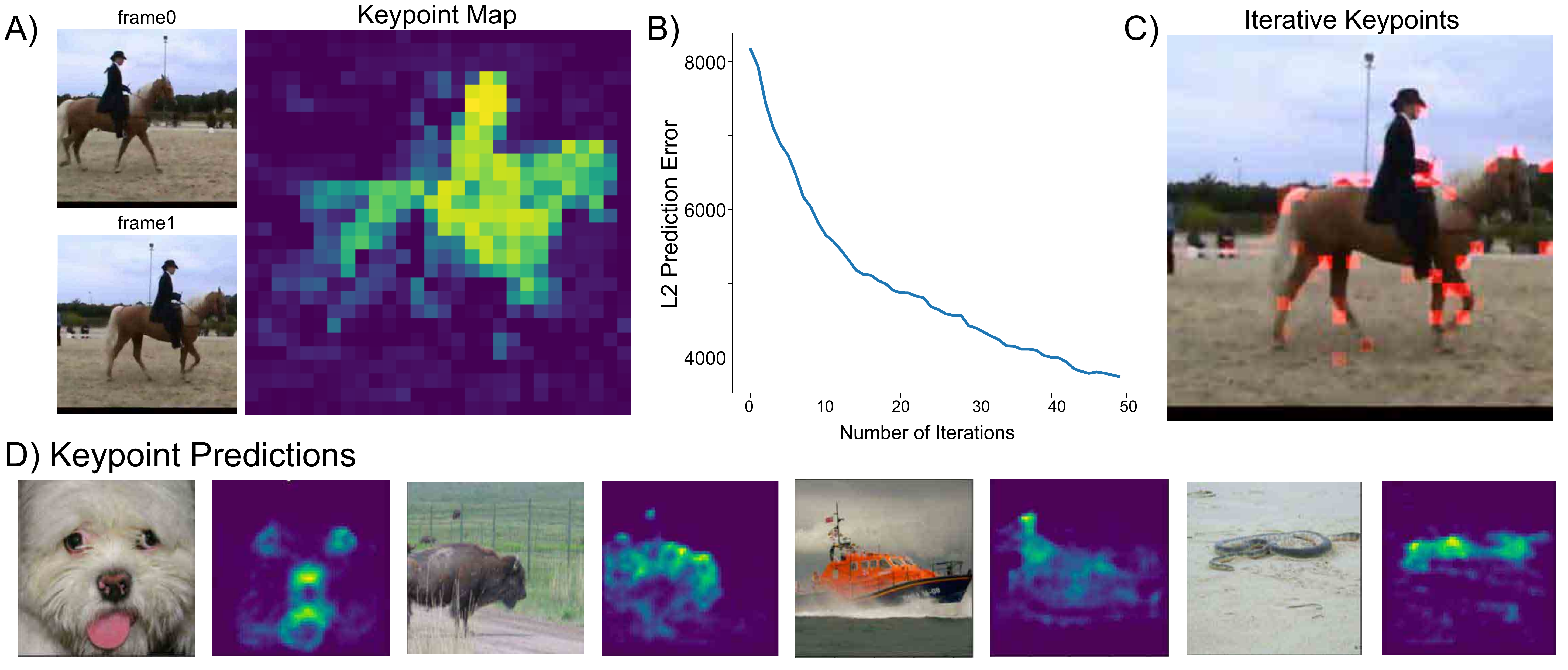}
\caption{
    \textbf{Keypoints.} \textbf{A)} For the indicated frame pair, the keypoint map shows the normalized error reduction achieved by revealing to the predictor $\Psi$ each single patch in \textbf{frame1} independently. For the same frame pair, \textbf{B)} shows the curve of prediction error (summed over pixels) as the iterative keypoint selection algorithm identifies keypoints, while \textbf{C)} shows the keypoints thereby selected. \textbf{D)} Consolidated keypoint maps as predicted from single frames.
}
\label{fig:keypoints}
\end{figure}

We compute keypoints in two ways: either by considering all potential visual patches at once (Fig. \ref{fig:keypoints}a) or by identifying them sequentially (Fig. \ref{fig:keypoints}b-c).
The first algorithm computes the \textit{reduction in total reconstruction error} upon revealing the full first frame and each second-frame patch, on its own, to $\Psi$.
This produces a map of each patch's ``importance'', with the most important patches taken to be keypoints of the scene.  
A second algorithm constructs keypoints iteratively, starting with an initial mask and adding visible tokens one-by-one to greedily reduce the reconstruction error. (See supplementary section \ref{s_sec:keypoints} for formal definitions of these constructions.) 
The mask sequences used in Fig.\ref{fig:tmp_predictions}a are constructed in the iterative fashion from $\Psi$, showing how semantically different subcomponents of the video naturally emerge in sequence as large, correlated portions of reconstruction error are reduced.

Computing keypoints with a masked predictor requires the ground-truth visual data on which to measure reconstruction error. 
However, given any masked prediction model $\Psi_g$, it is natural to formulate the problem of \textit{estimating} the keypoints of $\Psi_g$ from subsets of the patches made visible under masking policy $g$. Applied to $\Psi_{\textbf{TMP}}$, this boils down to learning which patches in a \textit{static} image would likely act as keypoints if that image were part of a video (Fig. \ref{fig:keypoints}d). Qualitatively, this recovers structures that are locally discriminative and invariant to (small) physical transformations -- the hallmarks of classical keypoints.

%% file: flow.tex
Since its proposal by J.J.Gibson~\cite{lee1980optic}, computation of \textit{optical flow} has been considered an essential component of object tracking, spatial navigation, forward prediction, and other behaviors in both animals and machine vision systems.
Optical flow is formally defined as the visual correspondence between elements in one frame of a video and elements in a nearby frame.
Classical approaches inferred these spatiotemporal correspondences by solving a regularized energy minimization problem, essentially trying to find matches between pixels in adjacent frames ~\cite{fleet2006optical}.  
More recent state-of-the-art methods supervise neural networks on ground truth optical flow from visually and physically realistic simulated data, using special-purpose architectures that favor energy minimization-like solutions~\cite{teed2020raft}.
Modern optical flow algorithms are often satisfactory for applications -- even on real-world movies outside their training distribution -- but do not unify this representation with other visual computations and do not explain how it could arise in an unsupervised way (e.g. in animals.)
Furthermore, even the best methods struggle to infer the motion of low-contrast regions that are intrinsically hard to track, which is addressed in practice via regularization but points to a missing deeper principle~\cite{stone2021smurf}.

\begin{figure}[t]
\centering
\includegraphics[width=\linewidth]{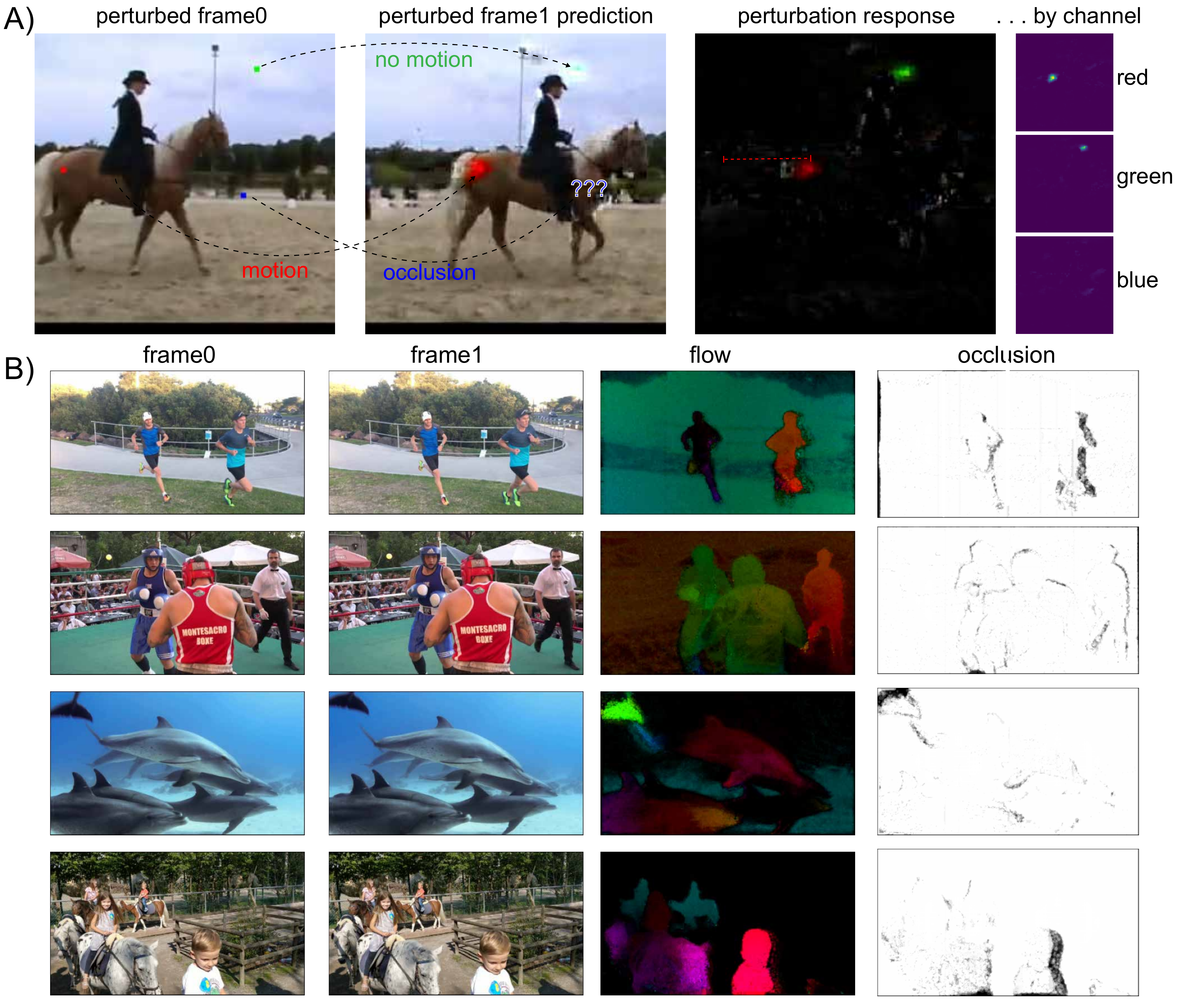}
\caption{
    \textbf{Flow extraction.}
    \textbf{A) Finite perturbation method.} Left Panel: Three small perturbations have been added to the original {\bf frame0}: a red patch on the horse's rear, a green patch on the sky, and a blue patch in front of the horse's nose.  
    Middle Panel: The perturbed input {\bf frame0} is then carried forward to a perturbed predicted next-frame by the predictor $\Psi$.  The red perturbation on the horse's rear has moved with the horse; the green perturbation in the sky has panned with the camera; and the blue perturbation has disappeared, having been occluded by the horse.   
    Right Panel: The perturbation response is the difference between $\Psi$'s prediction of the perturbed image and $\Psi$'s prediction of the original unperturbed ground ground-truth image.  This can be seen either in the composite RGB difference image (left) or in each color channel separately.  Dotted red line indicates displacement from original red perturbation in \textbf{frame0}.
    \textbf{B) Flows and Occlusions from Derivatives.} The method of derivatives allows for a tensor of infinitesimal counterfactual perturbations, analogous to those from the finite perturbation method, to be computed in parallel. The resulting flow and occlusion maps are shown for four example selections from the DAVIS dataset~\cite{pont20172017}. Note that, as non-simulated data, these videos have no ground-truth flow to compare to.
}
\label{fig:flow}
\end{figure}

In contrast, optical flow can be derived from the temporally-factored masked predictor $\Psi$ simply by ``asking it'' where counterfactual visual markers in frame $x_t$ would end up in frame $x_{t+\Delta}$ (Fig \ref{fig:flow}a).
Since $\Psi$ has learned to predict the dynamics of things in the first frame from small patches of the second frame, a marker placed on a visible surface should move to wherever $\Psi$ predicts that piece of surface is moving.

\begin{algorithm}[t]
\caption{Optical Flow via Appearance Counterfactuals 
\label{alg:flow}}
Let $(x_t, x_{t+1})$ be a pair of frames and $z \subset x_{t+1}$ a subset of patches in the second frame.
To estimate optical flow at location $\i$:
\begin{enumerate}
    \item Add a small perturbation to $x_t$ at location $\i$ to create a counterfactual first frame $$x_t^{\i} = x_t + \delta_\i.$$
    \item Fill in the rest of the second frame by applying the $\Psi$ to $x_t \oplus z$, which creates a ``clean'' prediction $$\hat{x}_{t+1} = \Psi(x_t \oplus z),$$ and to $x_t^{\i} \oplus z$, which creates a ``perturbed'' prediction: $$\hat{x}_{t+1}^{\i} = \Psi(x_t^{\i} \oplus z).$$
    \item Compute the \emph{perturbation response} -- the difference between the clean and perturbed predictions predictions, $$\hat{\delta}_{\i} = |\hat{x}^{\i}_{t+1} - \hat{x}_{t}|.$$ 
    \item Set the estimated optical flow at $\i$ to be the spatial displacement between the perturbed location and the peak (argmax over spatial positions) of the perturbation response $\hat{\delta}_{\i}$.
\end{enumerate}
\end{algorithm}

Below we formalize this counterfactual prompting procedure and explain how it can be considered, and implemented, as taking a derivative of the learned function $\Psi$.
Note that both in principle and in practice (Fig \ref{fig:flow}b), this approach does not require regularization or architectural inductive biases to compute optical flow on low-contrast regions: to the extent that $\Psi$ generalizes to the counterfactual input, it must properly transform elements of frame $x_t$ \textit{regardless of what they look like.}
Furthermore, the counterfactual method yields additional information compared to supervising on ground truth flow, as \textit{occlusions} are places where a counterfactual marker in frame $x_t$ appears to vanish in frame $x_{t+\Delta}$ (Fig \ref{fig:flow}a, blue point.)

\textbf{Optical Flow and Occlusion as Counterfactuals.} We can estimate optical flow using the algorithm shown in Algorithm box \ref{alg:flow}. Essentially, this is like adding a small ``tracer'' to the first frame at $\i$ that makes it (counterfactually) high-contrast, and therefore easy to localize in the second frame -- assuming $\Psi$ has enough information from $z$ to estimate the dynamics between $x_t$ and $x_{t+1}$.

This algorithm is simple and often effective (see Fig \ref{fig:flow}a) but can fail in two ways.
First, one of the revealed \textit{factual} patches in $z$ may cover the place where the perturbation at $\i$ is expected to move.
This is easily detected because $\hat{\delta}_{\i}$ will have small magnitude compared to $\delta_{\i}$.
It can be remedied by running the above algorithm for multiple choices of $z$ and taking their average result, removing samples where $|\hat{\delta}_p| / | \delta_{\i}| \ll 1$.
In the scenario where \textit{all} choices of $z$ lead to small-magnitude or poorly localized $\hat{\delta}_{\i}$, the content of $x_t$ at position $\i$ is likely occluded (or out of frame) in $x_{t+1}$.
Thus counterfactual prompting of the masked predictor $\Psi$ unifies estimation of optical flow and occlusion in a single algorithm.

\textbf{Counterfactuals as Derivatives of $\Psi$.} A second possible failure mode is that the perturbed frame $x_t^{\i}$ is out of distribution for $\Psi$. This could happen when the perturbation is either too large, so that it looks ``fake'' rather than like a piece of the surface it is placed on, or too small, so that it cannot be detected and moved accurately.
This is naturally addressed by considering $\Psi$'s \textit{infinitessimal responses to infinitessimal perturbations} -- which are exactly the \textit{derivative} of the prediction model:
\begin{align}
\textbf{Normalized Perturbation Response at $\i$} &= \lim_{\delta \i \rightarrow 0} \frac{|\hat{\delta}_{\i}|}{|\delta \i|}  \nonumber \\
  & = \lim_{\delta \i \rightarrow 0} \frac{|\hat{x}^{\i}_{t+1} - \hat{x}_{t+1}|}{|x^{\i}_t - x_t|} \nonumber \\
  & =  \nabla_{x}\Psi \bigg|_{\i}
\end{align}

To simultaneously estimate optical flow at all locations in an $H \times W$ frame pair, we can compute the Jacobian of $\Psi$.
This is a four-tensor $\J\Psi$ of shape $H \times W \times H \times W$, which assigns to element $(i, j, k, l)$ the predictor's change in output at location $(k,l)$ in the second frame due to an infinitessimal change at location $(i, j)$ in the first frame. Applying this derivative implementation to the algorithm above,
    \begin{equation}
        \textbf{flow}(k, l) = \begin{cases}
                   \textbf{undefined } \text{(disocclusion)} & \quad \text{if } \textbf{argmax}_{i,j}|\J\Psi(i,j,k,l)| \ll 1 \\
                   (k,l) - \textbf{argmax}_{i,j}\J\Psi(i,j,k,l)  & \quad \text{otherwise}
              \end{cases}
    \end{equation}
with results averaged over several choices of visible patches $z \subset x_{t+1}$.
(Note that this method detects disocclusion rather than occlusion, since no perturbation at any location in the first frame will cause a response at a point that becomes disoccluded in the second frame.)
Fig. \ref{fig:flow}b shows examples of flow computed this way for several image pairs. 

Because it is a tensorial operation, the derivative formulation of the counterfactual enables a more efficient parallel computation for flow than serial finite perturbations, implemented practically using Jacobian-vector products available in autograd packages such as PyTorch or Jax.\footnote{Speed can be further improved by consolidating the derivative-based flow readout as a self-supervised training signal for an efficient forward-inference predictor.}

%% file: segmentation.tex
Perceptual grouping of scenes into discrete objects is a fundamental computation in humans, where it forms a foundation for higher-level learning about object properties, dynamics, and relationships~\cite{todorovic2008gestalt}.
In computer vision the grouping task is formalized as \textit{object instance segmentation}.

While it is impossible to pin down general definitions of ``object'' or ``segmentation,'' an early developmental shift in perception suggests the practically useful notion of a \textit{Spelke object}.
Work by cognitive scientist Elizabeth Spelke and her colleagues revealed that babies begin life grouping separated visual elements into objects only when they are moving in concert~\cite{spelke1990principles}; it is not until they are almost a year old that babies expect, for example, two stationary toys resting on each other to be independently movable~\cite{carey2001infants}.
We therefore define a Spelke object as a \textit{collection of physical stuff that moves together during commonplace physical interactions}.
This includes common physical objects like cell phones, wallets, coffee cups, dogs, people, and cars, as well as complex novel objects from semantically unnamed categories that nonetheless can be seen to move coherently.
The ability to segment Spelke objects in an unsupervised, category-agnostic fashion is an open problem in computer vision and a clear \textit{desideratum} for robotic planning and control.

In previous work~\cite{chen2022unsupervised}, we found that optical flow in real-world videos can sometimes provide a strong training signal for segmenting \textit{static} objects;
this is the case when an entire object moves against a static background and is the only thing moving in the scene (except perhaps a well-known \textit{mover} object, such as a hand or robot arm, that can be ``explained away.'')
Most real-world scene dynamics are more complex, though, as they include (i) camera motion, (ii) multiple moving entities, and (iii) deformable objects whose parts may be moving by different amounts (or not at all.)

\begin{figure}[ht]
\centering
\includegraphics[width=\linewidth]{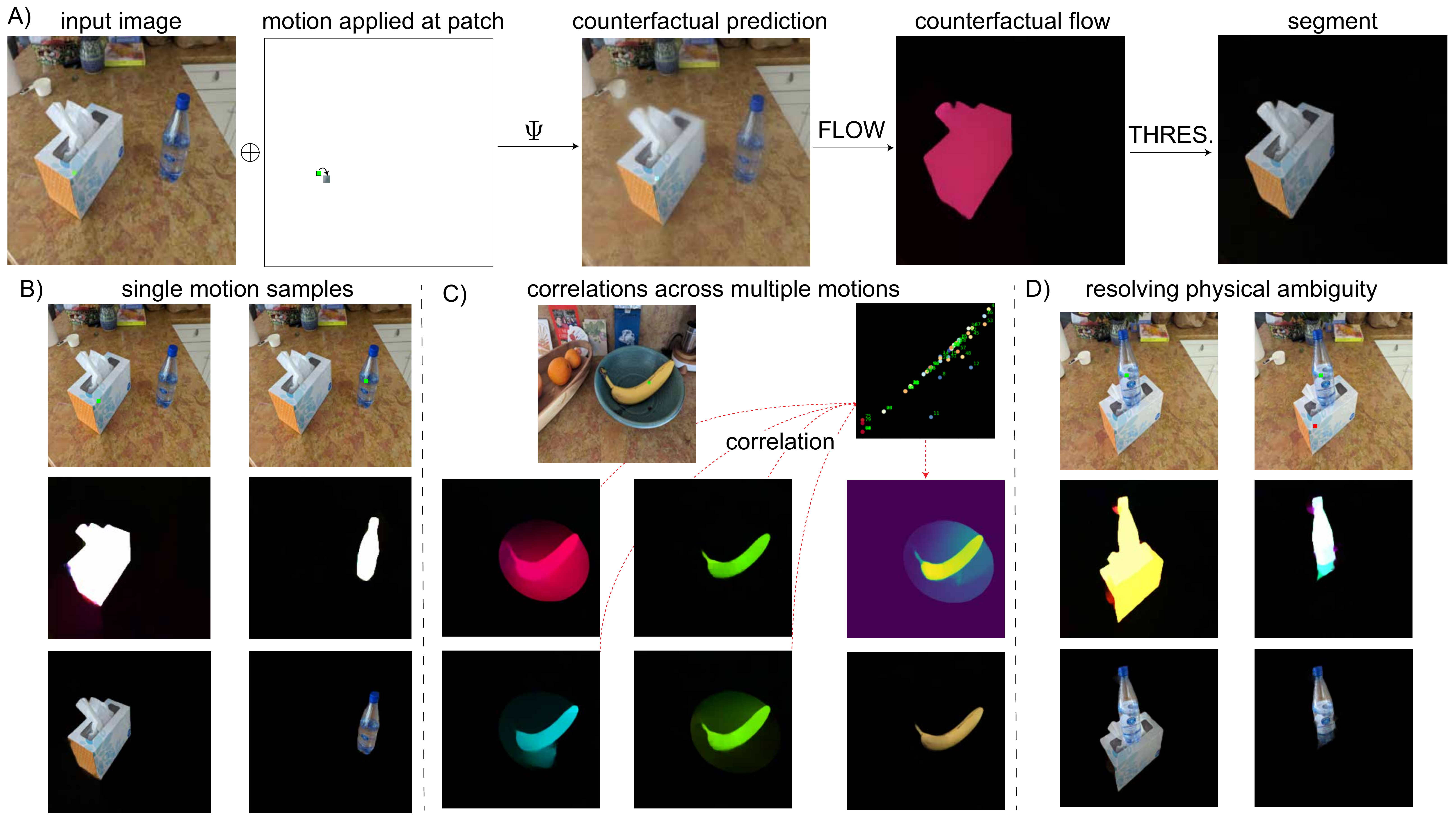}
\caption{
    \textbf{Movable-object segmentation.}
    \textbf{A) Segments from counterfactual motion.}  By placing a small amount of counterfactual motion at one location on an object (indicated by the small green square), a counterfactual prediction is generated in which the whole object moves.  The flow between the original input and the counterfactual prediction isolates the object.  \textbf{B)} shows two examples of this for isolated objects.  \textbf{C) Aggregating multiple motion samples.}  In more complex scenes, some motions may involve mutiple objects.  In this case, it is useful to collect multiple counterfactual motions and then compute motion affinities as the flow correlation over the counterfactuals. 
    \textbf{D) Resolving physical ambiguity with multi-point counterfactuals.}  Some situations -- such as one in which one object is on top of another -- will have unavoidable ambiguities in short-range motion correlations.  In this case, placing a single motion patch (green point) and a single stop-motion patch (red point) can disambiguate the proper Spelke object segment.
}
\label{fig:seg1}
\end{figure}

Counterfactual World Modeling with $\Psi$ and an optical flow estimator (see \ref{sec:flow} above) offers a natural way to resolve these issues.
Spelke objects can be extracted by asking not what \textit{is} moving in a real video (which might be complicated) but rather what \textit{could be} moving in a simulated video.
Specifically, motion counterfactuals (see Section \ref{sec:masked_predictors} for definition and Fig \ref{fig:tmp_predictions}b for examples) applied to a single frame $x$ produce estimates of what the optical flow would be if some patches were moving and others were static (Fig \ref{fig:seg1}a).
A formal algorithm for segmenting a Spelke object procedure is shown in Algorithm box \ref{alg:seg}.

\begin{algorithm}[t]
\caption{Movable-Object Segmentation via Motion Counterfactuals 
\label{alg:seg}}
\begin{enumerate}
    \item Pick a patch index $\i$.   
    \item Create the appearance of a counterfactual motion by taking the patch $x[p]$ at location $\i$ and creating a new frame that is largely blank, but in which the contents of the patch at location $\i$ has been copied (e.g. translated) to a new location $\i + \delta$. This creates a mostly-masked new frame $x'_\delta$.
    \item Apply $\Psi$ to the image pair consisting of the full original frame and the simulated masked motion frame, i.e. by computing $$\hat{x}_\delta = \Psi(x \oplus x'_\delta).$$  
    \item Now, compute the counterfactual flow $$f_\delta = \textbf{Flow}(x , \hat{x}_\delta).$$ 
    \item Declare the pixels $j$ at which $|f_\delta[j]| > 0$ to be part of the same Spelke object as that at location $\i$. 
\end{enumerate}
\end{algorithm}

\begin{figure}
\centering
\includegraphics[width=.9\linewidth]{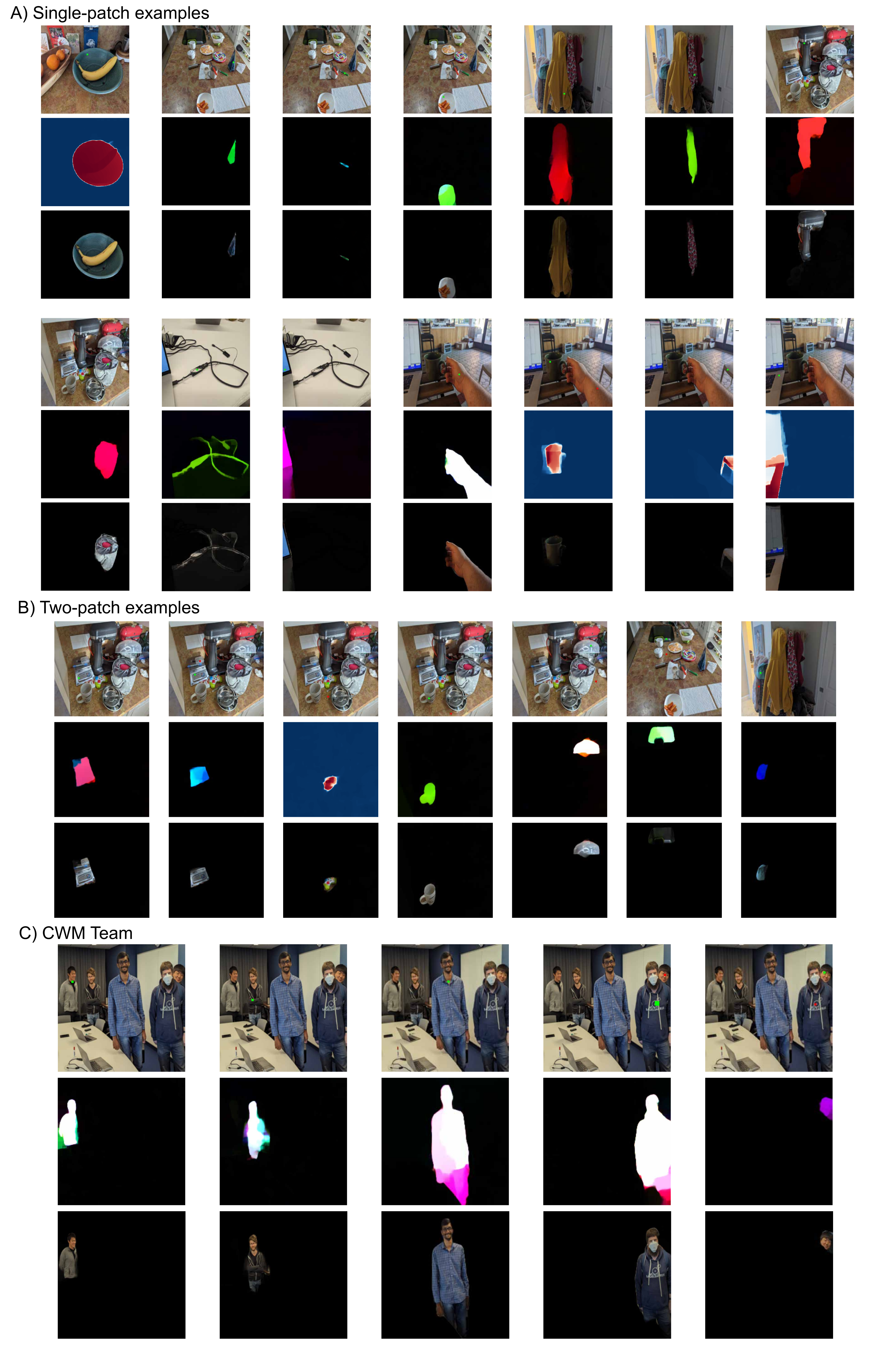}
\caption{
    \textbf{Some examples of segmentations.} 
    \textbf{A)} Examples where a single moving patch is sufficient to isolate Spelke objects. The selected patch is indicated by the small green square in each image.
    \textbf{B)} Examples where one moving patch (green square) and one stopped patch (red square) is sufficient.
    \textbf{C)} Isolating the people in a picture of (most of) the authorship team of this paper. 
\label{fig:seg2}
}
\end{figure}

The algorithm above works well for many objects in real-world scenes, as counterfactually moving a single patch causes the whole object to move (Fig \ref{fig:seg1}b).
This demonstrates that the notion of a Spelke object is learned implicitly by $\Psi$, without any segmentation labels: segmentation ability arises purely from having to predict which parts of a scene will move together under a wide variety of circumstances.
However, even this one conception of perceptual grouping contains a number of subtleties that must be addressed to arrive at a general-purpose segmenter; below we explain and show how more sophisticated Counterfactual World Modeling can deal with them.

\textbf{Spelke Affinities As Derivatives.}
As discussed above, many possible scene dynamics do not correspond to a single, entire objects moving on a static background.
Consider the limbs of an animal moving independently of its trunk, or one moving object colliding with another, for example.
Rather than thinking of coherent motion as all-or-none segments, it makes sense to estimate motion affinities~\cite{chen2022unsupervised} between pairs of scene elements, which can be low in one scene but high in another;
it is the set of \textit{possible} affinities -- that is, how often would two elements move together across a wide variety of dynamic scenes -- that supports a good representation of how objects should be grouped and how they can behave.

This relaxation of segments to pairwise affinities allows the above algorithm to be interpreted as another \textit{derivative} of $\Psi$ (combined with the optical flow estimator.)
We define the \textit{Spelke affinity} between two points $i$ and $j$ to be the response in the flow estimate at $j$ given a counterfactual change in the flow estimate at $i$. 
For any input frame pair $(x, y)$ (including frame pairs for which $x = y$, i.e. static images), in the infinitesimal limit of the motion counterfactual size, we can write the Spelke affinity between $i$ and $j$ as $\partial_{y_i}\textbf{Flow}(x, y)[j]$, or in tensorial form:
\begin{equation}
\textbf{SpelkeAffinity}[x] = \nabla_{y} \textbf{Flow}[x, y].
\end{equation}
Even when objects can undergo complex motion, such that pairs of points sometimes move together and sometimes do not, partial or complete Spelke affinities can be converted to (binarized) estimated segments via a differential grouping operation, Kaleidoscopic Propagation, introduced in~\cite{chen2022unsupervised}.
Besides providing an efficient method for segmenting Spelke objects, this derivative construction conceptually unifies segmentation with optical flow estimation -- illustrating that two apparently distinct computer vision tasks emerge, zero-shot, from the same unsupervised model.

\textbf{Accounting for Head Motion.}
A second complication for Spelke object segmentation is that apparent motion can be due to either a moving object or a moving camera.
Given a single-patch motion counterfactual, the base model $\Psi$ may infer apparent motion throughout the scene, as expected from a moving camera, and fail to isolate the target object.
We use the head-motion conditioned predictor $\Psi^{\textbf{H}}$ to solve this simply by \textit{also} conditioning on a (counterfactual) input of ``zero head motion.'' 
Optical flows from these multi-modal, joint counterfactual prompts are nearly always restricted to objects rather than backgrounds. 

\textbf{Averaging Multiple Motion Samples.}  
As with optical flow estimation, some choices of counterfactual may fail to yield a realistic prediction or a ``correct'' answer to the prompt.
For example, motion counterfactuals localized to low-contrast image patches, or those that would result in physically impossible motions, tend not to produce meaningful estimated segments (for good reason).
To ameliorate this, we assess multiple counterfactual motions of different magnitudes and directions across a variety of patch locations, then combine the results to define an aggregate Spelke affinity for any pair of locations $i, j$ (Fig. \ref{fig:seg1}c): i.e. $$\textbf{SpelkeAffinity}(i, j) \coloneqq \textbf{corr}(\vec{v}_i, \vec{v}_j)$$ where $\vec{v}_i$ (and similarly $\vec{v}_j$) is the vector of counterfactual flows $(|f_{\delta_0}[i]|, |f_{\delta_1}[i]|, \ldots)$ at location $i$, taken across multiple counterfactuals $\delta_0, \delta_1, \ldots$. 

\textbf{Resolving Intrinsic Uncertainty in Physical Dynamics.}
The remediation methods above produce more accurate Spelke object estimates, but often a single-patch motion counterfactual is still consistent with many possible scene dynamics; narrowing these down to isolate a target object is an intrinsic challenge for segmentation.
For example, inanimate objects do not typically move by themselves, but are instead moved be some effector (e.g. a hand). 
Similarly, if two objects are next to each other, moving one will often move the other, and thus produce motion in both (Fig. \ref{fig:seg1}d). 
Finally, the imposition of motion at just the patch $\i$ might be insufficient information for $\Psi$ to determine which of many possible motions to predict, yielding a counterfactual frame $\hat{x}_\delta$ that looks like a blurry average of different outcomes. 
Such images are out of the natural image distribution, and thus hard for the flow estimator to interpret. 

These uncertainty-related limitations -- which amount to failures of translating the definition of Spelke objects into a simple, single-patch motion counterfactual -- will be exacerbated to the extent that $\Psi$ is a \emph{correct} model of scene dynamics.  
But even though a single patch may be insufficient for segmenting some things, dynamics on short timescales are still highly constrained; 
as a result, only a few such patches are typically required to generate counterfactuals that isolate any object, even in complex scenes (Fig. \ref{fig:seg2}). 
Concretely, we can aim to select one or two additional co-moving patches, together with a similar number of patches deemed to be static (i.e. where no flow should be observed). 

With this goal in mind, we modify our procedure to sample two sets of patch indices $\textbf{Move} =\{\i_0, \i_1, \ldots\}$ and $\textbf{Stop} = \{\j_0, \j_1, \ldots, \}$ such that for each pair of patches $\i_1, \i_2 \in \textbf{Move}$ or $\i_1, \i_2 \in \textbf{Stop}$, the counterfactual motions generated for $\i_1$ and $\i_2$ have high mutual overlap; but for $\i \in \textbf{Move}$ and $\j \in \textbf{Stop}$, the mutual overlap between counterfactual motions is low. 
The procedure for determining the points with high affinity to $\i_0$ is then the same as before, but with the counterfactual generated by moving all the patches in $\textbf{Move}$ and fixing all those in $\textbf{Stop}$. 
Results of such a procedure can be see in Fig. \ref{fig:seg2}.   

%% file: depth.tex
\begin{algorithm}[t]
\caption{Relative depth via head-motion counterfactuals.
\label{alg:depth}}
\begin{enumerate}
\item Let $x$ be an image and let $\rho$ represent a head motion within the camera plane. 
\item Compute $x_{\rho}$ as the prediction of the $\Psi^{\textbf{H}}$ prompted only with $x$ and $\rho$, but no other information about the second video frame.
That is, $$x_{\rho} = \Psi^{\textbf{H}}(x \oplus \varnothing, \rho)$$ where $\varnothing$ represents the empty set of patches. 
\item Set the estimated relative depth, 
$$\textbf{RelativeDepth} = \frac{1}{\textbf{Flow}(x, x_{\rho})}.$$
\end{enumerate}
\end{algorithm}

\begin{figure}[t]
\centering
\includegraphics[width=\linewidth]{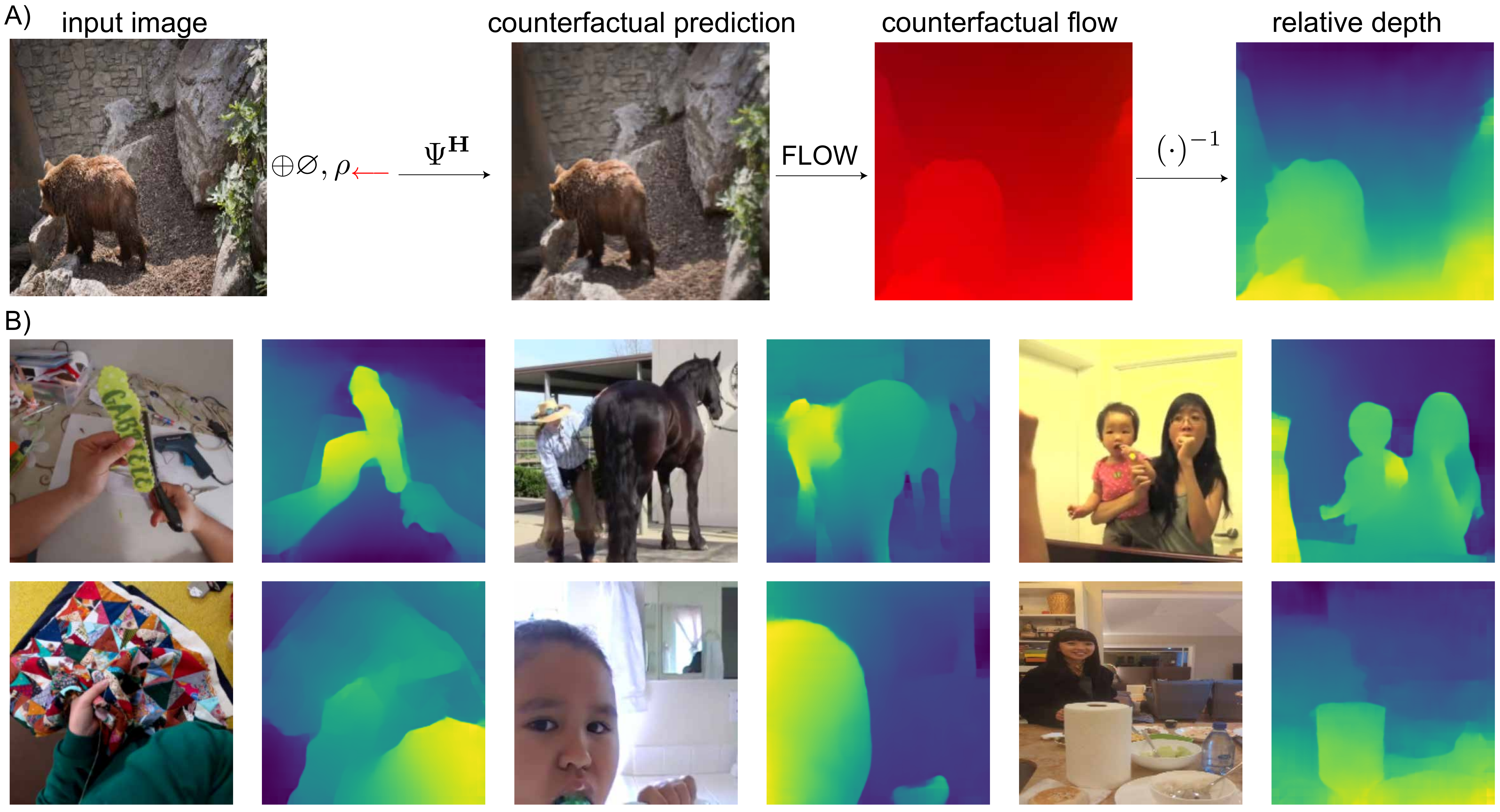}
\caption{
    \textbf{Estimating relative depth.}
    \textbf{A)} A process by which relative depth can be estimated from the head motion-conditioned predictor $\Psi^{\textbf{H}}$. Given an input image, a small within-plane camera-motion conditioning is applied, resulting in a counterfactual prediction. Relative depth is the reciprocal of the corresponding counterfactual flow; the latter is effectively a simulated disparity map for a small head motion.
    \textbf{B)} Some examples.
}
\label{fig:depth}
\end{figure}

As a final example of Counterfactual World Modeling, we illustrate how to estimate a relative depth map of an image by prompting a head-motion conditioned prediction model, $\Psi^{\textbf{H}}$.

In infants and animals, \textit{motion parallax} is a powerful cue for inferring the depth (distance) of elements of a scene~\cite{rogers1979motion}: when an observer translates laterally by a given amount, nearer things move across a greater portion of the visual field than farther things.
For small head motions and things in the center of the visual field, the relationship between apparent motion and depth is approximately inverse linear.\footnote{As can be seen by Taylor-expanding the right side of the relationship $\text{depth} = \frac{\rho}{\tan{\alpha}}$, where $\rho$ is the size of the small lateral head displacement and $\alpha$ the angular apparent motion of a visual element.}

This implies that relative depth could be estimated from a single, static image by \textit{simulating} the effect of a small, lateral head translation on apparent motion in the visual field -- that is, optical flow.
A simple counterfactual prompting algorithm is shown in Algorithm box \ref{alg:depth}.

Examples of depth estimates produced by this procedure are shown in Fig \ref{fig:depth}.
Note that, like optical flow and Spelke object segmentation, this algorithm could be implemented using derivatives rather than finite perturbations: in this case, relative depth would be the change in optical flow due to an infinitessimal in-camera-plane head motion. 
In practice we find that optical flow estimates with the RAFT architecture are better when counterfactually induced motion is $>1$ pixel.

This exercise shows how a single masked predictor carries implicit knowledge about tracked motion, segmentation, and 3D scene layout.
It is reasonable to wonder, however, whether counterfactually estimating depth in this way has any real-world value for computer vision or biological organisms.
After all, why simulate head motion when one can \textit{actually} move one's head (or have a second eye at the location one would move it to?)
Though speculative, our view is that running thought experiments about the contents of perception offers a way to reduce the effects of real-world observation noise and lean more strongly on an internal model of the world.
Elements of real scenes are not perfectly still while the head is moved, and one's eyes do not have the right separation to estimate depth at every distance range.
Thus an optimal system will need to find the right trade-off, for any given scene, between true action-dependent computations and estimates from its world model.

%% file: discussion.tex

In this work we have introduced the Counterfactual World Modeling framework, a way to produce a candidate pure-vision foundational model.  The CWM idea combines the observations that (i) specific statistical properties in a masked prediction problem force a model to learn a factoring of dynamical information from object-centric appearance information, and (ii) a wide variety of visual computations can be executed by counterfactual prediction with these models;
the latter can be implemented as taking (chained) derivatives of the model itself. 
We have shown that high-quality readouts of key visual concepts can be constructed, without any labeled data, via this form of zero-shot prompting.

The counterfactual physical reasoning abilities of CWM are a natural fit for applications that involve model-predictive control and planning, such as robotics~\cite{ebert2018visual}.
Though we have only shown the construction of ``2.5D'' visual properties in this work (keypoints, flow, occlusion, segmentation, and depth), it is natural to extend the approach here toward lifting multiview information to 3D structure, as evidenced by a variety of recent works~\cite{yu2021pixelnerf,wu2023multiview}.
We have postponed the question of how higher-level semantic structure (e.g. object categories) relate to our framework, an especially interesting topic given the recent success of contrastive unsupervised methods like DinoV2 in those domains~\cite{oquab2023dinov2}. 
Thought experiments with single-frame masked prediction hint at a CWM-based approach to categorization: for instance, an MAE that knows it should complete an elephant's body with a trunk, but not a rhino's, must have an implicit capacity to distinguish these two leathery, gray mammals.
Exploring these intuitions will be an important line of future work.

\begin{figure}[t]
\centering
\includegraphics[width=\linewidth]{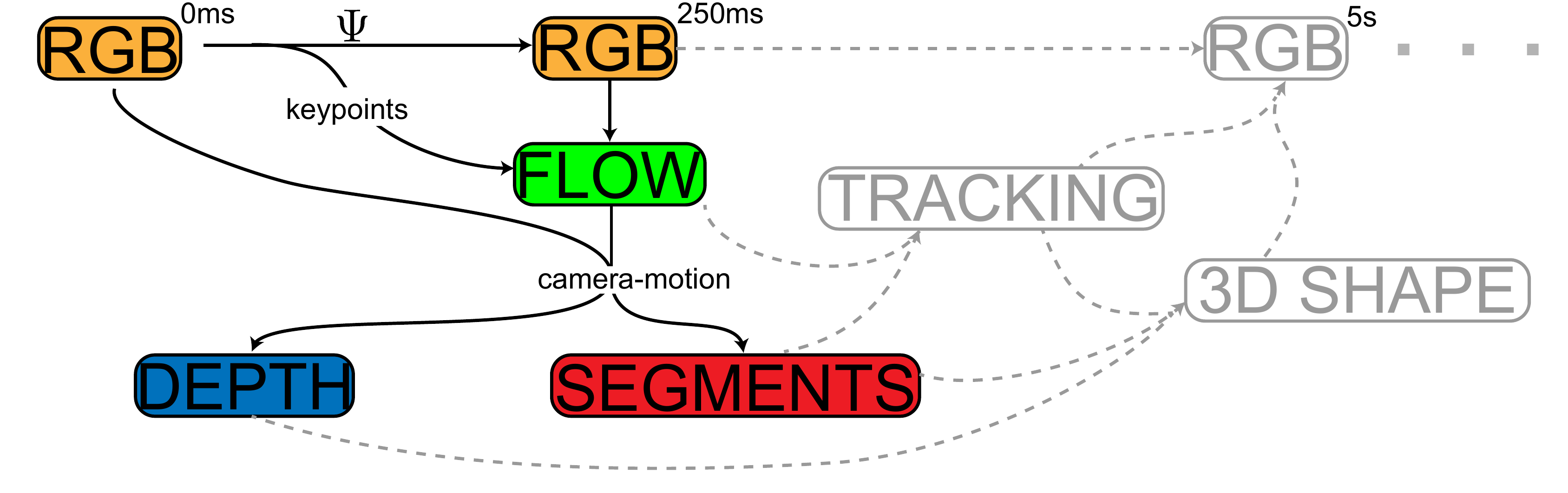}
\caption{
    \textbf{Counterfactual World Models as Computational Graphs.} The CWM predictor and its various readouts form an extensible computational graph, in which each component can be bootstrapped to construct further downstream task readouts. We hypothesize that the components illustrated in this paper (in color) will enable the bootstrapping of a variety of additional visual scene understanding tasks (in gray), including longer-range prediction. 
}
\label{fig:comp_graph}
\end{figure}

The model $\Psi$, like any masked predictor, is the approximation of a central tendency of a conditional generative model for the underlying data on which it is trained. It would be useful to leverage this observation to explicitly manage uncertainty in an improved version of CWM, in which not only is the central tendency of the masked tokens predicted, but also an estimate of their distribution's higher-order moments.  Doing so would likely resolve prediction failures that arise from mode collapse under the L2 pixel loss currently used to train $\Psi$ (see \ref{sec:segmentation}). 

A natural extension of CWM is to modalities other than vision, such as auditory or somatosensory data. 
While it is likely that the specific masking policies and counterfactuals that will be of use for extracting structure from (e.g.) auditory data are different from those useful in vision, the overall scheme of pairing a masked prediction model with counterfactual-based readouts may generalize.
The same may be true for multi-modal predictive models combining vision and language token streams. 

Another important next step in the development of the CWM framework will be to build longer-range prediction models. $\Psi$ operates effectively only at short timescales (< 1s). As we have shown, this turns out to be enough to extract a variety of important visual representations. It is possible that creating higher-level ''object tokens'' from the movable-object segments described in section \S\ref{sec:segmentation} would enable better ability to model complex physical interactions across multiple collisions~\cite{bear2021physion}. A natural hypothesis is that these intermediates could in turn be ``integrated'' as inputs to an improved model, functioning as useful inductive biases for bootstrapping to longer-range prediction (Fig. ~\ref{fig:comp_graph}).   

It would be especially useful if the process of discovering masking policies and useful derivative counterfactuals -- that is, building up the computational graph of CWM -- could be automated as a kind of evolutionary hyperparameter search~\cite{bergstra2013making,elsken2019neural}. It seems straightforward to parameterize the space of masking policies in terms of which fractions of which data slices are masked. The space of counterfactual token choices (or, equivalently, which chains of derivatives of the predictor are taken) may be similarly parameterized. We hypothesize that this hyperparameter space spans a wide spectrum of distinct but useful computational architectures, enabling efficient discovery of systems with the capability and universality required of true perceptual foundations.

\textbf{Acknowledgments.}
This work was supported by the following awards: To D.L.K.Y.: Simons Foundation grant 543061, National Science Foundation CAREER grant 1844724, Office of Naval Research grant S5122, ONR MURI 00010802 and ONR MURI S5847. D.M.B. is a Biogen Fellow of the Life Sciences Research Foundation and a Wu Tsai Neurosciences Institute Interdisciplinary Scholar. We also thank the Google TPU Research Cloud team for computing support. \emph{\textbf{S.D.G.}}

%% file: supplement.tex
\onecolumn 
\fancyhead{} 
\renewcommand{\floatpagefraction}{0.1}
\lfoot[\bSupInf]{\dAuthor}
\rfoot[\dAuthor]{\cSupInf}
\newpage

\captionsetup*{format=largeformat} 
\setcounter{figure}{0} 
\setcounter{equation}{0}
\setcounter{section}{0}
\makeatletter 
\renewcommand{\thefigure}{S\@arabic\c@figure} 
\makeatother
\def\theequation{S\arabic{equation}}
\renewcommand{\thesection}{\Alph{section}}

\newpage
\section{Technical Appendix}

\subsection{Formally Defining Masked Predictors}
\label{s_sec:formaldef}

We consider length-$\D$ video snippets, described as sequences of $H \times W$-pixel rectangular RGB frames $\x = (x_0, x_1, \ldots, x_{\D-1})$, and the tokens are rectangular $k \times l$-pixel patches tiling each frame. Let $\I_{\D}^{H,W,k,l}$ be the set of all indices of $k \times l$ patches in $H \times W \times \D$ movies (we will drop the superscript for ease of notation). For each index $i \in \I_{\D}$, let $\x[i]$ be the patch of $\x$ at index $i$, (e.g. the token value itself). For a subset of patches $p \subset \I$, let $\x[p] = \{(i, \x[i]) | i \in p\}$ -- the positionally-indexed set of token values in $p$ -- and let $\P_{\D}^{H, W, k, l}$ be set of all possible such token-value sets (again, we will typically drop the sub- and superscripts for readability). 

A \emph{mask generator} is a function $g$ that maps each $\x$ to a subset of $\I$, with the interpretation that a patch index $i$ is visible iff $i \in g(\x)$. For any mask generator $g$, let $\neg g(x)$ denote the complement of $g(x)$ in $\P(\x)$, i.e. the set of masked tokens.  For ease of notation, let $g[\x] = \x[g(\x)]$ and $\neg g[\x] = \x[\neg g(\x)]$, i.e. the sets of visible and masked tokens in $\x$ respectively.

The $g$-SMAE is then the predictor that tries to recover the masked tokens from the unmasked tokens.  
Formally, given a set of functions $F_{\th}: \P \longrightarrow \P$ with parameters $\th$ in some parameter space $\Th$, and a training dataset $\X$ of videos, let 
\begin{equation} \label{eq:smae}
    \th^*_{g,\X,\Th} = \argmin_{\th \in \Th}\ \E_{\X} [\L(F_{\Th}(g[x]), \neg g[x]) ],
\end{equation} where $\L(\cdot, \cdot)$ is a suitable loss function. The $g$-\emph{structured mask autoencoder on} $\X$ is then the function $F_{g,\X,\Th} \coloneqq F_{\th^*_{g,\X,\Th}}$. (We will drop the dependence on $\X$ and $\Th$ because these are usually clear from context.)  SMAEs are typically implemented by deep neural networks such as Vision Transformers (ViT)~\cite{} or Convolutional Neural Networks (CNNs)~\cite{}, and parameter optimization indicated in eq. \ref{eq:smae} is performed using standard gradient descent methods. Training datasets $\X$ are typically large natural image or video databases such as, e.g. ImageNet~\cite{} or Kinetics~\cite{}, and the loss function $\L$ is typically taken to be L1 or L2 pixel distance.

\subsection{Masked AutoEncoders and Video MAEs}

The Masked AutoEncoder (MAE)~\cite{he2022masked} is a masked predictor in which single RGB images are reconstructed from only a (randomly-chosen) fraction of each image given as input (Fig. \ref{fig:tmp}a). In this case, the temporal horizon $\D=1$, and for $p \in [0, 1]$, the mask generator is the random-valued function $g_p(\x)$ that uniformly randomly selects subsets of $\P(\x)$ of size $(1-p) \cdot k_H \cdot k_W$.

In the original MAE research, three basic facts were noticed.  First, unsurprisingly, performance on the reconstruction task decreases monotonically with $p$.  For large values of $p$ (e.g. $p>0.9$), performance is low -- since, presumably, having such a small portion of the input image simply doesn't provide the predictor with enough information to produce an effective guess of the output image. However, for any value of $p<\sim 0.8$, the reconstruction problem is well-solved by a sufficiently large network.  In other words, it is possible to learn a neural network that would complete real-world from a small number of patches. Second, the MAE turns out to be a good inpainter, with the ability to create effective background replacements the patches of an object are masked. Finally -- and perhaps most significantly -- as long as the value of $p$ is not too low, the features of the learned network support good transfer learning to tasks such as object categorization.  Essentially, as long as the MAE's prediction task is sufficiently non-trivial, the requirement to fill in images from a subset of the patches produces a good general-purpose image representation. Optimal results were obtained with $p \sim 0.75$. Many applications of the basic MAE have been made, across a wide variety of domains. 

Soon after the original MAE work, the idea was applied to video data in a direct fashion~\cite{tong2022videomae}. Various masking strategies are possible for videos, but the Video MAE (VMAE) work used a basic ``random tube'' strategy --- in which a random set of patches is chosen to be unmasked in the first frame, and the patches with the same spatial position (comprising ``temporal tubes'') are unmasked in all subsequent frames.  
The results of the VMAE were reminiscent of the original MAE, in that the directly-optimized reconstruction problem appeared to be reasonably well solved by large networks, and transfer to video tasks (e.g. action recognition) was possible.  

\subsection{Temporally-Factored Masked Predictors}
\label{s_sec:tmp}

The TMP is a predictor defined on length-2 video snippets $\x = (x_t, x_{t+\Delta})$, separated by a small temporal gap (e.g. $\Delta \sim$150ms).  The key ingredient for constructing the TMP is the mask generator in which \textbf{\emph{all} the patch tokens in the first frame, and only a very small number of patch tokens in the second frame, are given} (Fig: \ref{fig:tmp}b). This is formalized by the definition
\begin{equation}
\label{eq:tfmask}
g_{1p}(x_{t}, x_{t+\Delta}) = \I_1 \oplus g_p(x_{t+\Delta})
\end{equation} where $a \oplus b$ denotes the time-sequential concatenation of single-frame token index sets $a,b$, and $\I_1$ is short for $\I_1^{H,W,k,l}$.
Let $\TMP_p$ denote the predictor trained with this mask generator.
Unlike the typical MAE, where $p$ is chosen to be about 0.75, for the TMP, we choose $p$ to be much higher, e.g. $p > 0.99$.

\subsection{Keypoints}
\label{s_sec:keypoints}
Let $F$ be a trained masked predictor.  Suppose we are given an input $\x$, initial patch index set $p_0$, and an integer $k$.  The \emph{$k$-subsequent keypoint set for $\x$ from $p_0$} is
$$K_F(\x, p_0, k) = \argmin_{p\ \subset\ \I - p_0 \text{ such that } |p| = k} \L(F(\x_{p_0 \cup p}), \x).$$ 
Ideally it would be possible to efficiently compute $K_{F}(\x, p_0, k)$ directly for large $k$.  However, this is in general an intractable optimization problem.  Instead, it is substantially more efficient (though still rather inefficient in absolute terms) to take a greedy approach by finding one keypoint at a time and iterating, e.g. defining 
$$k_i = k_{i-1} \cup K_{F}(\x, p_0 \cup k_{i-1}, 1),\quad \quad \text{where } k_0 = \varnothing.$$

In the case of the TMP with two-frame input $(x_t, x_{t+\D})$, a natural starting point is to start with all patches in $x_t$ and no patches in $x_{t+\D}$, e.g  $p_0 = \I_1 \oplus \varnothing$.  This construction defines dynamical RGB keypoints.  Figure \ref{fig:keypoints} shows several examples of such keypoints for a variety of inputs, using the $\TMP_{0.01}$ model discussed above.

\textit{Keypoint consolidation.}  The keypoint construction is associated with several useful derived prediction tasks.  
One of these is the $\x$-conditioned sequence-to-sequence prediction task defined by
$$\textbf{Key}_{F}: (\x, p) \mapsto K_F(\x, p, k).$$  A predictor trained to solve this tasks can be thought of as \emph{consolidating} the keypoint construction into an efficient forward-inference procedure.  Also useful are the error and error difference prediction tasks defined respectively by:
$$\textbf{Err}_F: (\x, p) \mapsto \L(F(\x_p), \x)\quad \text{and} \quad \textbf{ErrDiff}_F: (\x, p, q) \mapsto \L(F(\x_p), \x) - \L(F(\x_q), \x).$$ 

\subsection{Head-Motion Conditioning}
\label{s_sec:head_motion}

In the Ego4D dataset~\cite{}, approximately 1/10th of the video data has usable IMU (inertial measurement unit) data obtained from the GoPro cameras used in data collection. To expand the use of this data, we first trained an IMU predictor from frame pairs $\textbf{H}(x_t, x_{t+\D}) \mapsto \text{IMU}(t-\D', t+\D)$ where $D' \sim 1850ms$, just on the subset of Ego4D data where reliable IMU is present. We then trained an IMU-conditioned TMP:
$$\TMPH_q: (x_t \oplus g_p[x_{t+\D}], h_t) \mapsto \neg g_p(x_{t+\D})$$ where $h_t = \textbf{H}(x_{t-D'}, x_{t+\D})$ is the estimated IMU value, on data from a union of Kinetics600 and Ego4D.

\subsection{Masking Variants}
\label{s_sec:mask_variants}

\begin{figure}[t]
\centering
\includegraphics[width=\linewidth]{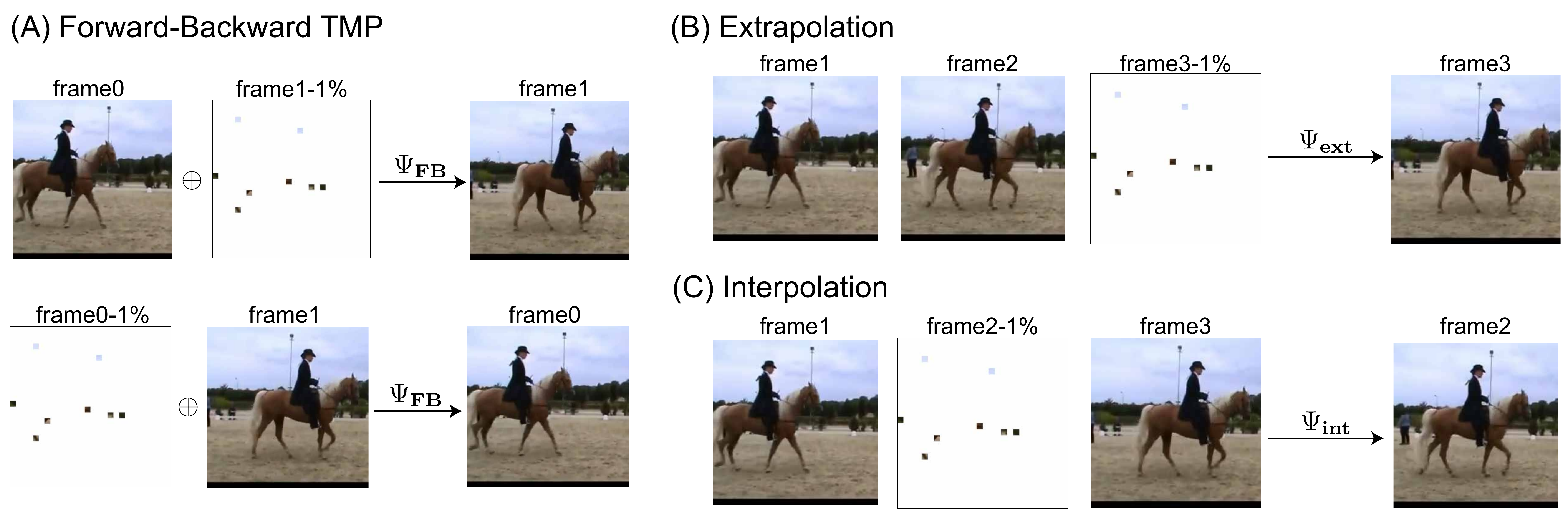}
\caption{
\textbf{Masking Policy Variants.}
\textbf{A)} The Forward-Backward TMP combines the vanilla TMP with an equivalent backward prediction (with 1\% of patches in \textbf{frame0} and all patches in \textbf{frame1}). The TMP-FB predictor is better than time-reversing the vanilla TMP at backward prediction for dynamics that are not time-reversal
symmetric. 
\textbf{B)} A similar approach allows the definition of interpolation and extrapolation predictors. 
}
\label{s_fig:alt_smae}
\end{figure}

The basic TMP idea can be generalized and extended in a variety of ways. 

\subsubsection*{Fractional Forward Masking} For example, the mask generator 
$$g_{qp}(x_{t}, x_{t+\Delta}) = g_q(x_t) \oplus g_p(x_{t+\D})$$
will train a predictor that has roughly the same second-frame reconstruction performance with $q=0.25$ as with $q=1$. This achieves a substantial efficiency in memory usage during training and speed at inference time.  Moreover, it endows the fully trained model with both dynamics-factoring properties of the TMP and the static object inpainting abilities of the regular MAE. 

\subsubsection*{Forward-Backward Masking}
Another useful variant of the masking policy is to encourage temporal factoring both forwards and backwards.  The mask generator
$$g^{\FB}_{1p}(\x) = \textbf{Bernoulli}_{0.5} \left(\{g_{1p}(\x), g_{p1}(\x)\} \right)$$
randomly selects between training forward prediction and backward predction.  This mask generator trains a predictor $\TMP^{\FB}_q$ that can predict forwards from $x_t$ to $x_{t+\D}$ or backwards from $x_{t+\Delta}$ to $x_t$ (Fig \ref{s_fig:alt_smae}a). Backward predictions from this model are better than ``time-reversed'' predictions of the original forward TMP, presumably because some aspects of natural motion are not statistically time-reversal symmetric (e.g. horses don't walk backwards by reversing their forward gait). 

\subsubsection*{Extrapolation and Interpolation}
More complex schemes of masking are possible.  For instance one can train predictors operating on 3 more frames, either using multiple context frames for forward extrapolation:
$$g^{\textbf{ext}}_{p}(x_t, x_{t+\D}, x_{t+2\D}) = \I_1 \oplus \I_1 \oplus g_p(x_{t+2\D})$$
or performing interpolation:
$$g^{\textbf{int}}_{p}(\x) = \I_1 \oplus g_p(x_{t+\D}) \oplus \I_1,$$
and a TMP with both interpolation and extrapolation properties is enabled simply by mixing the mask policies:
$$g^{\textbf{int-ext}}_{p}(\x) = \textbf{Bernoulli}_{0.5} \left(\{g^{\textbf{int}}_p(\x), g^{\textbf{ext}}_p(\x)\}\right).$$
More complex masking policies with varying combinations different fractional masking statistics, temporal patterns, and unions thereof, will generate predictors with more sophisticated scene and object understanding abilities.

%% file: wrapper.bbl
\begin{thebibliography}{49}
\providecommand{\natexlab}[1]{#1}
\providecommand{\url}[1]{\texttt{#1}}
\expandafter\ifx\csname urlstyle\endcsname\relax
  \providecommand{\doi}[1]{doi: #1}\else
  \providecommand{\doi}{doi: \begingroup \urlstyle{rm}\Url}\fi

\bibitem[Zhou et~al.(2000)Zhou, Friedman, and Von Der~Heydt]{zhou2000coding}
Hong Zhou, Howard~S Friedman, and R{\"u}diger Von Der~Heydt.
\newblock Coding of border ownership in monkey visual cortex.
\newblock \emph{Journal of Neuroscience}, 20\penalty0 (17):\penalty0
  6594--6611, 2000.

\bibitem[Beauchemin and Barron(1995)]{beauchemin1995computation}
Steven~S. Beauchemin and John~L. Barron.
\newblock The computation of optical flow.
\newblock \emph{ACM computing surveys (CSUR)}, 27\penalty0 (3):\penalty0
  433--466, 1995.

\bibitem[Lamme(1995)]{lamme1995neurophysiology}
V~A\_f Lamme.
\newblock The neurophysiology of figure-ground segregation in primary visual
  cortex.
\newblock \emph{Journal of neuroscience}, 15\penalty0 (2):\penalty0 1605--1615,
  1995.

\bibitem[Treisman(1982)]{treisman1982perceptual}
Anne Treisman.
\newblock Perceptual grouping and attention in visual search for features and
  for objects.
\newblock \emph{Journal of experimental psychology: human perception and
  performance}, 8\penalty0 (2):\penalty0 194, 1982.

\bibitem[Marr(2010)]{marr2010vision}
David Marr.
\newblock \emph{Vision: A computational investigation into the human
  representation and processing of visual information}.
\newblock MIT press, 2010.

\bibitem[Todd(2004)]{todd2004visual}
James~T Todd.
\newblock The visual perception of 3d shape.
\newblock \emph{Trends in cognitive sciences}, 8\penalty0 (3):\penalty0
  115--121, 2004.

\bibitem[Bear et~al.(2021)Bear, Wang, Mrowca, Binder, Tung, Pramod, Holdaway,
  Tao, Smith, Sun, et~al.]{bear2021physion}
Daniel~M Bear, Elias Wang, Damian Mrowca, Felix~J Binder, Hsiao-Yu~Fish Tung,
  RT~Pramod, Cameron Holdaway, Sirui Tao, Kevin Smith, Fan-Yun Sun, et~al.
\newblock Physion: Evaluating physical prediction from vision in humans and
  machines.
\newblock \emph{arXiv preprint arXiv:2106.08261}, 2021.

\bibitem[Rajalingham et~al.(2018)Rajalingham, Issa, Bashivan, Kar, Schmidt, and
  DiCarlo]{rajalingham2018large}
Rishi Rajalingham, Elias~B Issa, Pouya Bashivan, Kohitij Kar, Kailyn Schmidt,
  and James~J DiCarlo.
\newblock Large-scale, high-resolution comparison of the core visual object
  recognition behavior of humans, monkeys, and state-of-the-art deep artificial
  neural networks.
\newblock \emph{Journal of Neuroscience}, 38\penalty0 (33):\penalty0
  7255--7269, 2018.

\bibitem[Yamins et~al.(2014)Yamins, Hong, Cadieu, Solomon, Seibert, and
  DiCarlo]{yamins2014performance}
Daniel~LK Yamins, Ha~Hong, Charles~F Cadieu, Ethan~A Solomon, Darren Seibert,
  and James~J DiCarlo.
\newblock Performance-optimized hierarchical models predict neural responses in
  higher visual cortex.
\newblock \emph{Proceedings of the national academy of sciences}, 111\penalty0
  (23):\penalty0 8619--8624, 2014.

\bibitem[Zhuang et~al.(2021)Zhuang, Yan, Nayebi, Schrimpf, Frank, DiCarlo, and
  Yamins]{zhuang2021unsupervised}
Chengxu Zhuang, Siming Yan, Aran Nayebi, Martin Schrimpf, Michael~C Frank,
  James~J DiCarlo, and Daniel~LK Yamins.
\newblock Unsupervised neural network models of the ventral visual stream.
\newblock \emph{Proceedings of the National Academy of Sciences}, 118\penalty0
  (3):\penalty0 e2014196118, 2021.

\bibitem[Bommasani et~al.(2021)Bommasani, Hudson, Adeli, Altman, Arora, von
  Arx, Bernstein, Bohg, Bosselut, Brunskill,
  et~al.]{bommasani2021opportunities}
Rishi Bommasani, Drew~A Hudson, Ehsan Adeli, Russ Altman, Simran Arora, Sydney
  von Arx, Michael~S Bernstein, Jeannette Bohg, Antoine Bosselut, Emma
  Brunskill, et~al.
\newblock On the opportunities and risks of foundation models.
\newblock \emph{arXiv preprint arXiv:2108.07258}, 2021.

\bibitem[Radford et~al.(2019)Radford, Wu, Child, Luan, Amodei, Sutskever,
  et~al.]{radford2019language}
Alec Radford, Jeffrey Wu, Rewon Child, David Luan, Dario Amodei, Ilya
  Sutskever, et~al.
\newblock Language models are unsupervised multitask learners.
\newblock \emph{OpenAI blog}, 1\penalty0 (8):\penalty0 9, 2019.

\bibitem[Voleti et~al.(2022)Voleti, Jolicoeur-Martineau, and
  Pal]{voleti2022masked}
Vikram Voleti, Alexia Jolicoeur-Martineau, and Christopher Pal.
\newblock Masked conditional video diffusion for prediction, generation, and
  interpolation.
\newblock \emph{arXiv preprint arXiv:2205.09853}, 2022.

\bibitem[Yan et~al.(2022)Yan, Hafner, James, and Abbeel]{yan2022temporally}
Wilson Yan, Danijar Hafner, Stephen James, and Pieter Abbeel.
\newblock Temporally consistent video transformer for long-term video
  prediction.
\newblock \emph{arXiv preprint arXiv:2210.02396}, 2022.

\bibitem[Nair et~al.(2022)Nair, Rajeswaran, Kumar, Finn, and
  Gupta]{nair2022r3m}
Suraj Nair, Aravind Rajeswaran, Vikash Kumar, Chelsea Finn, and Abhinav Gupta.
\newblock R3m: A universal visual representation for robot manipulation.
\newblock \emph{arXiv preprint arXiv:2203.12601}, 2022.

\bibitem[Jaegle et~al.(2021)Jaegle, Borgeaud, Alayrac, Doersch, Ionescu, Ding,
  Koppula, Zoran, Brock, Shelhamer, et~al.]{jaegle2021perceiver}
Andrew Jaegle, Sebastian Borgeaud, Jean-Baptiste Alayrac, Carl Doersch, Catalin
  Ionescu, David Ding, Skanda Koppula, Daniel Zoran, Andrew Brock, Evan
  Shelhamer, et~al.
\newblock Perceiver io: A general architecture for structured inputs \&
  outputs.
\newblock \emph{arXiv preprint arXiv:2107.14795}, 2021.

\bibitem[Lu et~al.(2022)Lu, Clark, Zellers, Mottaghi, and
  Kembhavi]{lu2022unified}
Jiasen Lu, Christopher Clark, Rowan Zellers, Roozbeh Mottaghi, and Aniruddha
  Kembhavi.
\newblock Unified-io: A unified model for vision, language, and multi-modal
  tasks.
\newblock \emph{arXiv preprint arXiv:2206.08916}, 2022.

\bibitem[Chen et~al.(2022{\natexlab{a}})Chen, Saxena, Li, Lin, Fleet, and
  Hinton]{chen2022unified}
Ting Chen, Saurabh Saxena, Lala Li, Tsung-Yi Lin, David~J Fleet, and Geoffrey~E
  Hinton.
\newblock A unified sequence interface for vision tasks.
\newblock \emph{Advances in Neural Information Processing Systems},
  35:\penalty0 31333--31346, 2022{\natexlab{a}}.

\bibitem[Wu et~al.(2023{\natexlab{a}})Wu, Yin, Qi, Wang, Tang, and
  Duan]{wu2023visual}
Chenfei Wu, Shengming Yin, Weizhen Qi, Xiaodong Wang, Zecheng Tang, and Nan
  Duan.
\newblock Visual chatgpt: Talking, drawing and editing with visual foundation
  models.
\newblock \emph{arXiv preprint arXiv:2303.04671}, 2023{\natexlab{a}}.

\bibitem[Li et~al.(2023)Li, Zhu, Jiang, Zhu, Li, Yuan, Wang, Qiao, Wang, Wang,
  et~al.]{li2023uni}
Hao Li, Jinguo Zhu, Xiaohu Jiang, Xizhou Zhu, Hongsheng Li, Chun Yuan, Xiaohua
  Wang, Yu~Qiao, Xiaogang Wang, Wenhai Wang, et~al.
\newblock Uni-perceiver v2: A generalist model for large-scale vision and
  vision-language tasks.
\newblock In \emph{Proceedings of the IEEE/CVF Conference on Computer Vision
  and Pattern Recognition}, pages 2691--2700, 2023.

\bibitem[Wang et~al.(2023)Wang, Wang, Cao, Shen, and Huang]{wang2023images}
Xinlong Wang, Wen Wang, Yue Cao, Chunhua Shen, and Tiejun Huang.
\newblock Images speak in images: A generalist painter for in-context visual
  learning.
\newblock In \emph{Proceedings of the IEEE/CVF Conference on Computer Vision
  and Pattern Recognition}, pages 6830--6839, 2023.

\bibitem[He et~al.(2022)He, Chen, Xie, Li, Doll{\'a}r, and
  Girshick]{he2022masked}
Kaiming He, Xinlei Chen, Saining Xie, Yanghao Li, Piotr Doll{\'a}r, and Ross
  Girshick.
\newblock Masked autoencoders are scalable vision learners.
\newblock In \emph{Proceedings of the IEEE/CVF Conference on Computer Vision
  and Pattern Recognition}, pages 16000--16009, 2022.

\bibitem[Chen et~al.(2020)Chen, Kornblith, Norouzi, and Hinton]{chen2020simple}
Ting Chen, Simon Kornblith, Mohammad Norouzi, and Geoffrey Hinton.
\newblock A simple framework for contrastive learning of visual
  representations.
\newblock In \emph{International conference on machine learning}, pages
  1597--1607. PMLR, 2020.

\bibitem[Grill et~al.(2020)Grill, Strub, Altch{\'e}, Tallec, Richemond,
  Buchatskaya, Doersch, Avila~Pires, Guo, Gheshlaghi~Azar,
  et~al.]{grill2020bootstrap}
Jean-Bastien Grill, Florian Strub, Florent Altch{\'e}, Corentin Tallec, Pierre
  Richemond, Elena Buchatskaya, Carl Doersch, Bernardo Avila~Pires, Zhaohan
  Guo, Mohammad Gheshlaghi~Azar, et~al.
\newblock Bootstrap your own latent-a new approach to self-supervised learning.
\newblock \emph{Advances in neural information processing systems},
  33:\penalty0 21271--21284, 2020.

\bibitem[Kirillov et~al.(2023)Kirillov, Mintun, Ravi, Mao, Rolland, Gustafson,
  Xiao, Whitehead, Berg, Lo, et~al.]{kirillov2023segment}
Alexander Kirillov, Eric Mintun, Nikhila Ravi, Hanzi Mao, Chloe Rolland, Laura
  Gustafson, Tete Xiao, Spencer Whitehead, Alexander~C Berg, Wan-Yen Lo, et~al.
\newblock Segment anything.
\newblock \emph{arXiv preprint arXiv:2304.02643}, 2023.

\bibitem[Tong et~al.(2022)Tong, Song, Wang, and Wang]{tong2022videomae}
Zhan Tong, Yibing Song, Jue Wang, and Limin Wang.
\newblock Videomae: Masked autoencoders are data-efficient learners for
  self-supervised video pre-training.
\newblock \emph{arXiv preprint arXiv:2203.12602}, 2022.

\bibitem[Feichtenhofer et~al.(2022)Feichtenhofer, Fan, Li, and
  He]{feichtenhofer2022masked}
Christoph Feichtenhofer, Haoqi Fan, Yanghao Li, and Kaiming He.
\newblock Masked autoencoders as spatiotemporal learners, 2022.

\bibitem[Kay et~al.(2017)Kay, Carreira, Simonyan, Zhang, Hillier,
  Vijayanarasimhan, Viola, Green, Back, Natsev, et~al.]{kay2017kinetics}
Will Kay, Joao Carreira, Karen Simonyan, Brian Zhang, Chloe Hillier, Sudheendra
  Vijayanarasimhan, Fabio Viola, Tim Green, Trevor Back, Paul Natsev, et~al.
\newblock The kinetics human action video dataset.
\newblock \emph{arXiv preprint arXiv:1705.06950}, 2017.

\bibitem[Grauman et~al.(2022)Grauman, Westbury, Byrne, Chavis, Furnari,
  Girdhar, Hamburger, Jiang, Liu, Liu, et~al.]{grauman2022ego4d}
Kristen Grauman, Andrew Westbury, Eugene Byrne, Zachary Chavis, Antonino
  Furnari, Rohit Girdhar, Jackson Hamburger, Hao Jiang, Miao Liu, Xingyu Liu,
  et~al.
\newblock Ego4d: Around the world in 3,000 hours of egocentric video.
\newblock In \emph{Proceedings of the IEEE/CVF Conference on Computer Vision
  and Pattern Recognition}, pages 18995--19012, 2022.

\bibitem[Day and Fitzpatrick(2005)]{day2005vestibular}
Brian~L Day and Richard~C Fitzpatrick.
\newblock The vestibular system.
\newblock \emph{Current biology}, 15\penalty0 (15):\penalty0 R583--R586, 2005.

\bibitem[Crapse and Sommer(2008)]{crapse2008corollary}
Trinity~B Crapse and Marc~A Sommer.
\newblock Corollary discharge across the animal kingdom.
\newblock \emph{Nature Reviews Neuroscience}, 9\penalty0 (8):\penalty0
  587--600, 2008.

\bibitem[Minderer et~al.(2019)Minderer, Sun, Villegas, Cole, Murphy, and
  Lee]{minderer2019unsupervised}
Matthias Minderer, Chen Sun, Ruben Villegas, Forrester Cole, Kevin~P Murphy,
  and Honglak Lee.
\newblock Unsupervised learning of object structure and dynamics from videos.
\newblock \emph{Advances in Neural Information Processing Systems}, 32, 2019.

\bibitem[Mian et~al.(2008)Mian, Bennamoun, and Owens]{mian2008keypoint}
Ajmal~S Mian, Mohammed Bennamoun, and Robyn Owens.
\newblock Keypoint detection and local feature matching for textured 3d face
  recognition.
\newblock \emph{International Journal of Computer Vision}, 79:\penalty0 1--12,
  2008.

\bibitem[Lee(1980)]{lee1980optic}
Denis~N Lee.
\newblock The optic flow field: The foundation of vision.
\newblock \emph{Philosophical Transactions of the Royal Society of London. B,
  Biological Sciences}, 290\penalty0 (1038):\penalty0 169--179, 1980.

\bibitem[Fleet and Weiss(2006)]{fleet2006optical}
David Fleet and Yair Weiss.
\newblock Optical flow estimation.
\newblock \emph{Handbook of mathematical models in computer vision}, pages
  237--257, 2006.

\bibitem[Teed and Deng(2020)]{teed2020raft}
Zachary Teed and Jia Deng.
\newblock Raft: Recurrent all-pairs field transforms for optical flow.
\newblock In \emph{Computer Vision--ECCV 2020: 16th European Conference,
  Glasgow, UK, August 23--28, 2020, Proceedings, Part II 16}, pages 402--419.
  Springer, 2020.

\bibitem[Stone et~al.(2021)Stone, Maurer, Ayvaci, Angelova, and
  Jonschkowski]{stone2021smurf}
Austin Stone, Daniel Maurer, Alper Ayvaci, Anelia Angelova, and Rico
  Jonschkowski.
\newblock Smurf: Self-teaching multi-frame unsupervised raft with full-image
  warping.
\newblock In \emph{Proceedings of the IEEE/CVF Conference on Computer Vision
  and Pattern Recognition}, pages 3887--3896, 2021.

\bibitem[Pont-Tuset et~al.(2017)Pont-Tuset, Perazzi, Caelles, Arbel{\'a}ez,
  Sorkine-Hornung, and Van~Gool]{pont20172017}
Jordi Pont-Tuset, Federico Perazzi, Sergi Caelles, Pablo Arbel{\'a}ez, Alex
  Sorkine-Hornung, and Luc Van~Gool.
\newblock The 2017 davis challenge on video object segmentation.
\newblock \emph{arXiv preprint arXiv:1704.00675}, 2017.

\bibitem[Todorovic(2008)]{todorovic2008gestalt}
Dejan Todorovic.
\newblock Gestalt principles.
\newblock \emph{Scholarpedia}, 3\penalty0 (12):\penalty0 5345, 2008.

\bibitem[Spelke(1990)]{spelke1990principles}
Elizabeth~S Spelke.
\newblock Principles of object perception.
\newblock \emph{Cognitive science}, 14\penalty0 (1):\penalty0 29--56, 1990.

\bibitem[Carey and Xu(2001)]{carey2001infants}
Susan Carey and Fei Xu.
\newblock Infants' knowledge of objects: Beyond object files and object
  tracking.
\newblock \emph{Cognition}, 80\penalty0 (1-2):\penalty0 179--213, 2001.

\bibitem[Chen et~al.(2022{\natexlab{b}})Chen, Venkatesh, Friedman, Wu,
  Tenenbaum, Yamins, and Bear]{chen2022unsupervised}
Honglin Chen, Rahul Venkatesh, Yoni Friedman, Jiajun Wu, Joshua~B Tenenbaum,
  Daniel~LK Yamins, and Daniel~M Bear.
\newblock Unsupervised segmentation in real-world images via spelke object
  inference.
\newblock In \emph{Computer Vision--ECCV 2022: 17th European Conference, Tel
  Aviv, Israel, October 23--27, 2022, Proceedings, Part XXIX}, pages 719--735.
  Springer, 2022{\natexlab{b}}.

\bibitem[Rogers and Graham(1979)]{rogers1979motion}
Brian Rogers and Maureen Graham.
\newblock Motion parallax as an independent cue for depth perception.
\newblock \emph{Perception}, 8\penalty0 (2):\penalty0 125--134, 1979.

\bibitem[Ebert et~al.(2018)Ebert, Finn, Dasari, Xie, Lee, and
  Levine]{ebert2018visual}
Frederik Ebert, Chelsea Finn, Sudeep Dasari, Annie Xie, Alex Lee, and Sergey
  Levine.
\newblock Visual foresight: Model-based deep reinforcement learning for
  vision-based robotic control.
\newblock \emph{arXiv preprint arXiv:1812.00568}, 2018.

\bibitem[Yu et~al.(2021)Yu, Ye, Tancik, and Kanazawa]{yu2021pixelnerf}
Alex Yu, Vickie Ye, Matthew Tancik, and Angjoo Kanazawa.
\newblock pixelnerf: Neural radiance fields from one or few images.
\newblock In \emph{Proceedings of the IEEE/CVF Conference on Computer Vision
  and Pattern Recognition}, pages 4578--4587, 2021.

\bibitem[Wu et~al.(2023{\natexlab{b}})Wu, Johnson, Malik, Feichtenhofer, and
  Gkioxari]{wu2023multiview}
Chao-Yuan Wu, Justin Johnson, Jitendra Malik, Christoph Feichtenhofer, and
  Georgia Gkioxari.
\newblock Multiview compressive coding for 3d reconstruction.
\newblock \emph{arXiv preprint arXiv:2301.08247}, 2023{\natexlab{b}}.

\bibitem[Oquab et~al.(2023)Oquab, Darcet, Moutakanni, Vo, Szafraniec, Khalidov,
  Fernandez, Haziza, Massa, El-Nouby, et~al.]{oquab2023dinov2}
Maxime Oquab, Timoth{\'e}e Darcet, Th{\'e}o Moutakanni, Huy Vo, Marc
  Szafraniec, Vasil Khalidov, Pierre Fernandez, Daniel Haziza, Francisco Massa,
  Alaaeldin El-Nouby, et~al.
\newblock Dinov2: Learning robust visual features without supervision.
\newblock \emph{arXiv preprint arXiv:2304.07193}, 2023.

\bibitem[Bergstra et~al.(2013)Bergstra, Yamins, and Cox]{bergstra2013making}
James Bergstra, Daniel Yamins, and David Cox.
\newblock Making a science of model search: Hyperparameter optimization in
  hundreds of dimensions for vision architectures.
\newblock In \emph{International conference on machine learning}, pages
  115--123. PMLR, 2013.

\bibitem[Elsken et~al.(2019)Elsken, Metzen, and Hutter]{elsken2019neural}
Thomas Elsken, Jan~Hendrik Metzen, and Frank Hutter.
\newblock Neural architecture search: A survey.
\newblock \emph{The Journal of Machine Learning Research}, 20\penalty0
  (1):\penalty0 1997--2017, 2019.

\end{thebibliography}
